\documentclass[10pt,journal]{IEEEtran}
\usepackage{cite}

\usepackage{multirow}
\usepackage{multicol}
\usepackage{amsmath,graphicx,url}
\usepackage{tikz-cd,pgfplots,pgfgantt,pdflscape}
\usetikzlibrary{arrows}
\usepackage{tikz-3dplot}
\usepackage{amssymb,amsthm}
\pgfplotsset{compat=1.18}
\usepackage[ruled,vlined]{algorithm2e}
\DontPrintSemicolon
\SetAlgoNlRelativeSize{0}
\SetAlgoNlRelativeSize{-1}

\newtheorem{theorem}{Theorem}
\newtheorem{assumption}{Assumption}
\newtheorem{lemma}{Lemma}
\newtheorem{remark}{Remark}

\title{Exponentially Consistent Nonparametric Linkage-Based Clustering of Data Sequences}

\author{Bhupender Singh$^{\dagger}$\quad Ananth Ram Rajagopalan$^{\star}$\quad Srikrishna Bhashyam$^{\dagger}$

\vspace{0.4 cm}

$^{\dagger}$ Indian Institute of Technology Madras, Chennai 600036, India \\ $^{\star}$ Purdue University, West Lafayette, IN 47907, USA \\ bhupenderiitm@gmail.com, r.ananthram14@gmail.com,  skrishna@ee.iitm.ac.in

\thanks{This work was done at the Department of Electrical Engineering, Indian Institute of Technology Madras, Chennai, India. Part of this work was presented at IEEE ICASSP 2024, Seoul, Korea \cite{10446356}.}}

\allowdisplaybreaks

\begin{document}
\IEEEtitleabstractindextext{%

\begin{abstract}
In this paper, we consider nonparametric clustering of $M$ independent and identically distributed (i.i.d.) data sequences generated from {\em unknown} distributions. The distributions of the $M$ data sequences belong to $K$ underlying distribution clusters. Existing results on exponentially consistent nonparametric clustering algorithms, like single linkage-based (SLINK) clustering and $k$-medoids distribution clustering, assume that the maximum intra-cluster distance ($d_L$) is smaller than the minimum inter-cluster distance ($d_H$). First, in the fixed sample size (FSS) setting, we show that exponential consistency can be achieved for SLINK clustering under a less strict assumption, $d_I < d_H$, where $d_I$ is the maximum distance between any two sub-clusters of a cluster that partition the cluster. Note that $d_I < d_L$ in general. Thus,  our results show that SLINK is exponentially consistent for a larger class of problems than previously known. In our simulations, we also identify examples where $k$-medoids clustering is unable to find the true clusters, but SLINK is exponentially consistent. Then, we propose a sequential clustering algorithm, named SLINK-SEQ, based on SLINK and prove that it is also exponentially consistent. Simulation results show that the SLINK-SEQ algorithm requires fewer expected number of samples than the FSS SLINK algorithm for the same probability of error. 
\end{abstract}

}

\maketitle

\IEEEdisplaynontitleabstractindextext
\section{Introduction}
\label{sec:intro}

We consider the problem of clustering independent and identically distributed (i.i.d.) data sequences generated from {\em unknown} probability distributions under the nonparametric setting. In other words, each data sequence is a sequence of i.i.d. samples generated from an unknown distribution. The data sequences have to be clustered according to the closeness of the underlying distributions. Since the distributions are unknown and not modelled using any parametric distribution class, our algorithms are nonparametric and universal. Clustering of data sequences has applications in several practical problems like market segmentation \cite{marketsegdoi:10.1137/1.9781611972788.58,JMLR:v25:22-1088}, image retrieval \cite{imageretLIN20143276}, clustering variants of viruses \cite{JMLR:v25:22-1088}. The special case of anomalous sequence detection has several applications including network intrusion detection and fraud detection \cite{outlying6879597}.  

Algorithms to solve the general problem of clustering a set of data points \{$ X^{(1)}, X^{(2)},....., X^{(M)}$\}, with each data point  $ X^{(i)}$ $\in \mathbb{R}^n $ have been well studied, see for example  \cite{JMLR:v6:banerjee05b,KauRou87,gower1969minimum,johnson1967hierarchical,shalev-shwartz_ben-david_2014,huang2018hierarchical,bishop2006pattern,oyewole2023data}. These algorithms could be classified into two categories: partitional (e.g. $k$-means \cite{JMLR:v6:banerjee05b}, $k$-medoids \cite{KauRou87}) and hierarchical \cite{huang2018hierarchical} (e.g. single-linkage (SLINK) \cite{gower1969minimum}, complete-linkage (CLINK)  \cite{johnson1967hierarchical}). The focus of these works was on the computational complexity of the clustering algorithms. Data stream clustering is studied in \cite{10.1145/2522968.2522981,10.1007/s10462-020-09874-x,9646912,8734059,8736821} where observations arrive in a continuous stream and these observations have to be clustered in an online manner under constraints on storing/reading past observations. The data stream could also evolve in a non-stationary manner. However, this problem is different from the clustering of data sequences problem that we study in this paper. In our problem, multiple data sequences are observed and the sequences are clustered, i.e., each data sequence is assigned to a single cluster. 

Clustering of i.i.d. data sequences has been studied under two settings: fixed sample-size (FSS) and sequential (SEQ). In the FSS setting, we have $n$ samples from each data sequence and the clustering algorithm groups the $M$ data sequences into $K$ clusters. The clustering algorithm output is denoted as $\{C_1(n),\hdots,C_{K}(n)\}$, where $C_i(n)$ is the set of indices of data sequences in the $i^{th}$ output cluster. When the clustering output does not match the true underlying clusters denoted $\{P_1,\hdots,P_K\}$, there is an error, i.e., the error event is given by  $E =\{ \{C_1(n),\hdots,C_{K}(n) \}  \neq   \{P_1,\hdots,P_K\}\}$. The clustering algorithm is consistent if the probability of error $P_e = \mathbb{P}[E] \rightarrow 0$ as $n \rightarrow \infty$, and is exponentially consistent if $P_e < a e^{-bn}$ for some positive $a, b$ and sufficiently large $n$. In the SEQ setting, we get one sample from each data sequence at a each time and the clustering algorithm decides whether to output the clusters and stop or continue to get more samples. In this case, consistency and exponential consistency are defined based on the relationship between probability of error and the expected number of samples taken by the algorithm before it stops. For the FSS setting,  \cite{8651294} analyzed the error probability of the $k$-medoids clustering algorithm and proved exponential consistency.
For this setting, distance metrics between distributions, e.g. maximum mean discrepancy (MMD), and Kolmogorov-Smirnov distance (KSD), are more suitable than Euclidean distance between the $X^{(i)}$'s \cite{JMLR:v13:gretton12a,sriperumbudur2012empirical,8651294}. The exponential consistency of linkage-based hierarchical clustering algorithms was studied in \cite{wang2020exponentially}. The performance of anomaly detection, a special case of clustering, was analyzed in the nonparametric setting in \cite{8651359,doi:10.1080/07474946.2017.1360086,sreenivasan2021sequential}.

In \cite{8651294}, exponential consistency of $k$-medoids clustering was proved in the fixed sample-size (FSS) setting under the assumption that the maximum intra-cluster distance ($d_L$) is smaller than the minimum inter-cluster distance ($d_H$). A similar result was later proved for the $k$-medoids clustering algorithm in the sequential setting in \cite{sreenivasan2023nonparametric}. In \cite{wang2020exponentially}, linkage-based hierarchical clustering algorithms (including the SLINK algorithm) were also shown to be exponentially consistent in the FSS setting under the same assumption $d_L < d_H$. However, settings where $d_L > d_H$ are also important. For example, we could have clusters that have large diameter, but are separated by distances that are smaller than the diameter: see  \cite{NEURIPS2019_9b16759a}, where a clustering problem with two such clusters is considered. In the clustering problem in \cite{NEURIPS2019_9b16759a}, images are clustered into groups based on a safety score to assess the safety of a given scene.

In this paper, we make the following contributions: 
\begin{itemize}
    \item We first show that single-linkage clustering (SLINK) in the FSS setting is exponentially consistent under a less strict assumption than $d_L < d_H$ for the underlying distribution clusters. We define $d_I$ to be the maximum distance between any two sub-clusters of a cluster that partition the cluster and require the assumption that $d_I < d_H$. Note that $d_I < d_L$ when there are more than two distinct points in a cluster. 
    \item We identify and simulate numerical examples where SLINK is exponentially consistent while $k$-medoids clustering is not. Thus, we identify that different clustering algorithms require different conditions on the separation of underlying distribution clusters.
    \item Then, we develop a nonparametric sequential clustering algorithm, SLINK-SEQ, based on SLINK. We show that SLINK-SEQ is also exponentially consistent. 
    \item Using simulation results on synthetic and real data sets, we show that the proposed sequential SLINK clustering outperforms FSS SLINK clustering in terms of expected number of samples required for a given probability of error.
\end{itemize}  
{Overall, we demonstrate that, in the FSS setting, SLINK clustering of data sequences can achieve exponential consistency under a less restrictive condition than previously known, and subsequently, propose an exponentially consistent clustering algorithm, SLINK-SEQ, in the sequential setting.}

The rest of the paper is organized as follows. In Section \ref{sec:model}, we provide the problem statement and the required notation. In Section \ref{sec:expSLINK}, we prove the exponential consistency of SLINK clustering under the assumption $d_I < d_H$. Then, we propose the nonparametric sequential SLINK clustering algorithm in Section \ref{sec:seqSLINK} and prove that it is exponentially consistent. Simulation results and conclusions are presented in Sections \ref{sec:simulations} and \ref{sec:conclusions}, respectively.

\begin{table}
\caption{Important Parameters and Notation}
    \begin{tabular}{|c|p{6cm}|} 
        \hline
        \textbf{Symbol} & \textbf{Description} \\
        \hline
        $M$ & Number of data sequences to be clustered \\ \hline
        $K$ & Number of clusters \\ \hline
        $\{C_i(n)\}$ & Clustering output using $n$ samples from each sequence\\
        \hline
        $\{P_i\}$ & True clusters  \\ \hline
        $N$ & Stopping time of clustering algorithm   \\ \hline
        ${\mathbb{E}}[N]$ & Expected number of samples taken before stopping  \\
        \hline
        $P_e$ & Probability of Error \\ \hline
        $d_{\text{ks}}(p, q)$ & KS distance between distributions $p$ and $q$ \\
        \hline
        $d_{\text{mmd}}(p,q)$ & MMD distance between distributions $p$ and $q$  \\
        \hline
        $d(p, q)$ & Distance between distributions $p$ and $q$ \\ \hline
        $d_L$ & Maximum intra-cluster distance \\
        \hline
        $d_H$ & Minimum inter-cluster distance \\
        \hline
        $d_I$ & Maximum distance between any
two sub-clusters of a cluster that partition the cluster. \\
        \hline
        $KS(i,j,n)$ & Estimate of KS distance between sequences $i$ and $j$ using $n$ samples\\
        \hline
        $M_b(i,j,n)$ & Estimate of MMD distance between sequences $i$ and $j$ using $n$ samples\\
        \hline
        $\hat{d}(i,j,n)$ & Estimate of distance between sequences $i$ and $j$ using $n$ samples\\
        \hline
        $k(x, y)$ & Kernel function used to estimate MMD \\ \hline
        ${\mathbb{G}}$ & Upper bound on the kernel function\\
        \hline
        $C$ & Parameter in threshold of SLINK-SEQ algorithm \\ \hline 
    \end{tabular}
    \vspace{0.3 cm}
    
    \label{tab:notation}
    \end{table}

\section{System Model and Preliminaries}
\label{sec:model}
\subsection{Clustering problem setup}
Let $M$ distinct data sequences denoted \{$X^{(1)},X^{(2)},\hdots,X^{(M)}$\} be observed. Each of these sequences is an i.i.d. sequence of samples from an unknown distribution. These distributions belong to one of $K$ distribution clusters represented as $\{D_{1},D_{2}, \hdots,D_{K}\}$. Each distribution cluster $D_k$ is comprised of $M_k$ distributions, i.e., $D_k = \{p_{k}^{j} : j = 1,2,\hdots,M_k\}$ for each  $k = 1,2,\hdots,K$. 
$P_k$ denotes the set of indices of data sequences that are drawn from distributions in the  $k^{th}$ distribution cluster, i.e., $P_k = \{i : X^{(i)} \sim p_{k}^{j},~ j = 1,2,\hdots,M_k\}$. 

First, we consider the FSS setting where $n$ samples are observed for each data sequence. The clustering algorithm output is denoted as $\{C_1(n),\hdots,C_{K}(n)\}$, where $C_i(n)$ is the set of indices of data sequences in the $i^{th}$ output cluster.  The error event is the event that the clustering output $\{C_1(n),\hdots,C_{K}(n)\}$ does not match the true underlying clusters denoted $\{P_1,\hdots,P_K\}$, and is denoted  $E =\{ \{C_1(n),\hdots,C_{K}(n) \}  \neq   \{P_1,\hdots,P_K\}\}$.

Then, we consider the sequential clustering setting in Section \ref{sec:seqSLINK}. Here, at each time $n$, one new sample is observed in each of the $M$ data sequences. We denote the stopping time of the algorithm by $N$ and the final clustering output by $\{C_1(N),\hdots,C_{K}(N)\}$. Here, the error event $E =\{ \{C_1(N),\hdots,C_{K}(N) \}  \neq   \{P_1,\hdots,P_K\}\}$. The important parameters and notation are summarized in Table \ref{tab:notation}.

\subsection{Consistency and Exponential Consistency}
{\em FSS setting:} An FSS clustering algorithm is consistent if the probability of error $P_e = \mathbb{P}[E] \rightarrow 0$ as $n \rightarrow \infty$. The clustering algorithm is exponentially consistent if $P_e < a e^{-bn}$ for some positive $a, b$ and sufficiently large $n$, or, equivalently, \cite{8651359}
\[
\lim_{n \rightarrow \infty} -\frac{1}{n} \ln P_e = b > 0.
\]

{\em SEQ setting:} A sequential clustering algorithm is consistent if $P_e$ goes to zero as $E[N]$ goes to infinity, and is exponentially consistent if the probability of error decreases exponentially with increasing $E[N]$, or, equivalently, \cite{doi:10.1080/07474946.2017.1360086}
\[
    \mathbb{E}[N] \le \frac{-\log P_{e}}{\alpha} (1 + o(1)),
\]
with $\alpha > 0$.

Since we do not make any assumptions about the underlying distributions for each data sequence, our algorithms are {\em universal}. Therefore, we also refer to the consistency property as universal consistency. The term nonparametric also refers to the same property, i.e., we do not assume any parametric model or family of distributions for the distributions of the data sequences.

\subsection{Distances between data sequences}
Let $d(p,q)$ denote the distance between two i.i.d. data sequences with samples from distributions $p$ and $q$, respectively.\footnote{Either Maximum mean discrepancy (MMD) or Kolmogorov-Smirnov distance (KSD) is used in this paper.} 
As in \cite{8651294}, we consider the two distances, MMD and KSD, in this work to calculate the distance. Since these distances can be estimated well from the observed data sequences without any information about the distributions, our clustering algorithms are universal. 
The KSD is given by 
\[d_{\text{KS}}(p, q) = \sup_{a \in \mathbb{R}} |F_p(a) - F_q(a)|,\]
where $F(\cdot)$ denotes the CDF. The MMD is given by
\[d_{\text{MMD}}(p, q) = \sup_{f \in {\mathcal{F}}} \left( E_p [ f(X) ] - E_q [ f(Y) ] \right),\]
where ${\mathcal{F}}$ is chosen as in \cite{JMLR:v13:gretton12a}. This choice ensures that the MMD is a metric and also has computable estimates. The estimates of KSD (denoted $KS(i, j, n)$) and MMD (denoted $M_b(i,j,n)$) between sequences $i$ and $j$ using $n$ observed samples for each sequence are as follows \cite{8651294,sreenivasan2023nonparametric}: 
\begin{equation}
    KS(i, j, n) = \sup_{a \in \mathbb{R}} |\hat{F}_i^{(n)}(a) - \hat{F}_j^{(n)}(a)|,
    \label{eqn:ksdest}
\end{equation}
where 
\[\hat{F}_i^{(n)}(a) = \frac{1}{n} \sum_{l=1}^{n} I_{[-\infty,a]}(X^{(i)}_l),\]
and 
\begin{equation}
    M_b(i,j,n) ={\frac{1}{n^2}\sum_{l,m=1}^{n}h(X_l^{(i)},X_m^{(i)},X_l^{(j)},X_m^{(j)})},
    \label{eqn:mmdest}
\end{equation}
where 
\[h(X_l^{(i)},X_m^{(i)},X_l^{(j)},X_m^{(j)})=\hspace{1.5 in} \]
\[ k(X_l^{(i)},X_m^{(i)}) + k(X_l^{(j)},X_m^{(j)}) - 2k(X_l^{(i)},X_m^{(j)}),\]
and $k(x,y)$ is a kernel function, with $0 \le k(x,y) \le \mathbb{G}$ for all $x, y$, i.e, $\mathbb{G}$ is the upper bound on the kernel function. In the rest of the paper, we refer to the MMD or KSD estimates using the notation $\hat{d}(i,j,n)$.

\begin{figure}[!ht]
    \centering
    \includegraphics[width=0.5\textwidth]{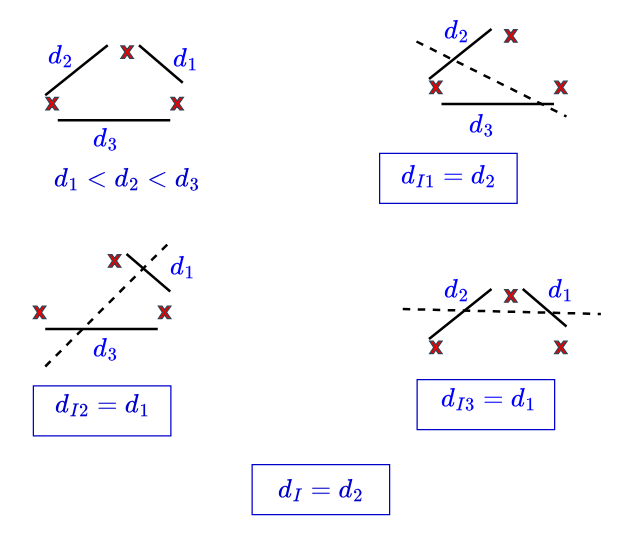}
    \caption{Illustration of $d_I$ for a cluster with 3 distributions. The pairwise distances are $d_1, d_2$ and $d_3$ with $d_1 < d_2 < d_3$. The three ways to partition this cluster into 2 are shown. The corresponding minimum distance between these sub-clusters is also indicated: $d_2$ in one case and $d_1$ in the other cases. Overall, the $d_I$ for this cluster is $d_2$. The overall $d_I$ for the problem will be the maximum of the $d_I$'s over all the clusters. Note that the distances are actually the distances between the distributions and not Euclidean distance.}
    \label{fig:dIcluster}
\end{figure}
\subsection{Cluster distances: $d_L$, $d_H$ and $d_I$}

We first define the following distances: minimum inter-cluster distance $d_H$, and maximum intra-cluster distance $d_L$.\footnote{These definitions for $d_L$ and $d_H$ are the same as in \cite{8651294}.} The inter-cluster distance between clusters $D_{k}$ and  $D_{l}$ is given by:
\[ d(D_k, D_l) = \min_{p \in D_k, q \in D_l} d(p, q).\]
The minimum inter-cluster distance $d_H$ is the minimum over all cluster pairs and is given by:
\begin{equation}
d_H = \min_{k \neq l} d(D_k, D_{l}).
\label{eqn:dH}
\end{equation}
The maximum intra-cluster distance (diameter) of cluster $D_{k}$ is:
\[ d(D_k) = \max_{p \in D_k} \max_{q \in D_k} d(p, q). \]
The overall maximum intra-cluster distance is the maximum of the above distance over all the clusters, i.e., 
\[d_L = \max_{k=1,\hdots,K} d(D_k).\]
Now, we define a new quantity $d_I(D_k)$ for each cluster $D_k$. This is the maximum distance between two sub-clusters of  cluster $D_k$ that partition $D_k$. Thus, for cluster $k$, $d_I(D_k)$ is given by 
\begin{equation}
    d_I(D_k) = \max _{\substack{D_{1k}, D_{2k} \\ D_{1k} \cup D_{2k} = D_k \\ D_{1k} \bigcap D_{2k} = \phi}} \min _{\substack{p, q \\ p \in D_{1k} \\ q \in D_{2k}}} d(p, q),
\label{eqn:didefn}
\end{equation}
In (\ref{eqn:didefn}), the inner minimum gives the minimum distance between two sub-clusters $D_{1k}$ and $D_{2k}$. These two sub-clusters partition $D_k$, i.e., $D_{1k} \bigcap D_{2k} = \phi$ and $D_{1k} \cup D_{2k} = D_k$. The outer maximum gives the maximum over all such partitions. An illustration is provided in Fig. \ref{fig:dIcluster} for a cluster with 3 distributions. Finally, we define $d_I$ as the maximum over all clusters of $d_I(D_k)$:
\begin{equation}
d_I = \max_{k=1,\hdots,K} d_I(D_k).
\label{eqn:dI}
\end{equation}
We can also write $d_I(D_k)$ in terms of $P_k$ (which contains the indices of the data sequences in cluster $k$) as follows.
\[d_I(D_k) = \max _{\substack{P_{1k}, P_{2k} \\ P_{1k} \cup P_{2k} = P_k \\ P_{1k} \bigcap P_{2k} = \phi}} \min _{\substack{i, j \\ i \in P_{1k} \\ i \in P_{2k}}} d(i, j),\]
Here, $i$ and $j$ in $d(i,j)$ are data sequence indices. Note that $d_I(D_k) < d(D_k)$ when there are more than two distinct points in cluster $k$, and hence $d_I < d_L$, in general.\footnote{In the conference paper \cite{10446356}, we considered the maximum nearest neighbour distance as $d_I$, i.e., only partitions with only one element in one of the sub-clusters. However, we also needed the clustering to be unique under the given $d_I < d_H$ condition. The improved definition in (\ref{eqn:didefn}) provides the unique clustering inherently and does not require any extra condition other than $d_I < d_H$.}

In this paper, we focus on problems where the underlying distribution clusters $\{D_{1},D_{2}, \hdots,D_{K}\}$ satisfy $d_I < d_H$. In particular, 
in our consistency and exponential consistency analysis, we assume $d_I < d_H$, which is less strict compared to $d_L < d_H$ assumed in \cite{8651294,wang2020exponentially,sreenivasan2023nonparametric}. Some examples that illustrate this difference are presented later in the simulation results section.

\section{Exponential consistency of SLINK in the fixed sample size setting}
\label{sec:expSLINK}
In this section, we prove the exponential consistency of single linkage-based (SLINK) clustering in the FSS setting, also referred to as SLINK FSS,  under the assumption that $d_I < d_H$. As mentioned earlier, this assumption is less strict than the assumption of $d_L < d_H$ used in the analysis of SLINK in \cite{wang2020exponentially,wangthesis}.

\subsection{SLINK clustering of data sequences}

SLINK can be summarized as follows. In the FSS setting, we have $n$ samples of each data sequence. First, the pairwise distance between each pair of data sequences is estimated (using MMD or KSD estimates). Then, the clustering of data sequences is carried out as follows. Initially, each data sequence is considered to be a cluster, i.e., we start with $M$ clusters. Then, we find the two closest clusters, where the inter-cluster distance is the minimum distance between any data sequence in one cluster from any data sequence in the other cluster. These two clusters are merged. The algorithm stops when the number of clusters is $K$ (assuming $K$ is known). When the number of clusters $K$ is unknown, the algorithm continues as long as the two closest clusters are closer than $d_H$ and stops if the distance between the two closest clusters is greater than or equal to $d_H$.\footnote{Either the knowledge of the number of clusters, or the knowledge of the expected minimum distance between two clusters is required. A lower bound on $d_H$ would also suffice.}

\subsection{SLINK with true distances}
\label{sec:slinktrue}
If the exact pairwise distances between the distributions of the different data sequences are provided as input to the SLINK algorithm, it can be argued in the following manner that the clustering output will be correct (under the assumption $d_I < d_H$). From the definition of $d_I$, we know that any sub-cluster of cluster $k$ is within $d_I$ of at least one other sequence in the same cluster $k$. From the definition of $d_H$, we know that for any sub-cluster of cluster $k$ the nearest neighbour from any other cluster is farther than $d_H$. Therefore, due to the assumption that $d_I < d_H$, during each step of SLINK sub-clusters belonging to the same true cluster will be chosen for merging till all $K$ clusters have been identified, i.e., there will be no error if the true pairwise distances are known. 

In practice, since the true distributions are assumed to be unknown, the SLINK algorithm is provided with the estimates of the pairwise MMD or KSD distances between the data sequences. In the following section, we analyze the performance of SLINK with estimated distances and prove exponential consistency.

\subsection{SLINK consistency with estimated MMD}
Now, we show that SLINK is exponentially consistent if $d_I < d_H$. We show the proof for the algorithm with MMD estimates. A similar proof can be provided even for the case with KSD based on the concentration results for KSD in \cite{8651294}. 
\begin{theorem}
   Let $d_I < d_H$. The probability of error of SLINK is upper bounded as: $P_e \le a_f e^{-b_f n}$ for some $a_f, b_f > 0$ for sufficiently large $n$.
   \label{thm:fssconsistency}
\end{theorem}

\begin{proof}
First, we let $\hat{d}_I$ and $\hat{d}_H$ denote the $d_I$ and $d_H$ corresponding to the true clusters using the estimated distances,\footnote{Note that $\hat{d}_I$ and $\hat{d}_H$ are only defined for the purpose of the analysis and are not actually estimated. We use the definitions in (\ref{eqn:dH}) and (\ref{eqn:dI}) to determine $\hat{d}_H$ and $\hat{d}_I$, respectively.} i.e., we have \[\hat{d}_I = \max_k\max_{\substack{P_{1k}, P_{2k} \\ P_{1k} \cup P_{2k} = P_k \\ P_{1k} \bigcap P_{2k} = \phi}} \left[\min_{i \in P_{1k}, j \in P_{2k}}\hat{d}(i,j,n)\right],\]
\[\hat{d}_H =\min_{k, k', k \ne k'}[\min_{i \in P_k}\min_{j \in P_{k'}}\hat{d}(i,j,n)].\]
Now, we identify that if $\hat{d}_I < \hat{d}_H$, then there will be no error in clustering (as discussed Sec. \ref{sec:slinktrue}). Therefore, we have
\[
P_e \le 1 - \mathbb{P}[\hat{d}_I < \hat{d}_H].
\]
In the rest of the proof, we show that $1 - \mathbb{P}[\hat{d}_I < \hat{d}_H]$ decreases exponentially in $n$ for large $n$.
Let $d_{th}$ be such that $d_I < d_{th} < d_H$. Now, we can write:
\begin{eqnarray}
    \mathbb{P}[\hat{d}_I < \hat{d}_H] &= \mathbb{P}[\hat{d}_I < \hat{d}_H | \hat{d}_H > d_{th}] ~\mathbb{P}[\hat{d}_H > d_{th}]\nonumber \\
&+ \mathbb{P}[\hat{d}_I < \hat{d}_H | \hat{d}_H \le d_{th}] ~\mathbb{P}[\hat{d}_H \le d_{th}]\nonumber \\
&\ge \mathbb{P}[\hat{d}_I < \hat{d}_H | \hat{d}_H > d_{th}] ~\mathbb{P}[\hat{d}_H > d_{th}]\nonumber \\
&\ge \mathbb{P}[\hat{d}_I < d_{th} | \hat{d}_H > d_{th}] ~\mathbb{P}[\hat{d}_H > d_{th}]\nonumber \\
&= \mathbb{P}[\hat{d}_I < d_{th}] ~\mathbb{P}[\hat{d}_H > d_{th}]. \nonumber \hspace{0.5in}
\end{eqnarray}
Now, we individually bound $\mathbb{P}[\hat{d}_H < d_{th}]$ and $\mathbb{P}[\hat{d}_I > d_{th}]$. For each bound, we rely on the convergence of MMD estimates to the true MMD. For bounding $\mathbb{P}[\hat{d}_I < d_{th}]$, we also use the definition of $d_I$ carefully while using the union bound. 
\begin{align*}
\mathbb{P}\left(\hat{d}_H<d_{t h}\right) &= \mathbb{P}\left[\bigcup_{\substack{k,k' \\ k \ne k'}} \bigcup_{\substack{i \in P_k \\ j \in P_{k'}}} \left\{ \hat{d}(i, j, n)<d_{th} \right\} \right]  \\
&\leq M^2 \mathbb{P}[\hat{d}(i, j, n)<d_{th}]  \\
&\text{(where $i$, $j$ are from different clusters)} \\
&\leq M^2\left[2 \exp \left(\frac{-n\left(d_H-d_{th}\right)^2}{16 \mathbb{G}}\right)\right] \\
&= a_H e^{-b_H n},
\end{align*}
where $a_H = 2 M^2$ and $b_H = \frac{\left(d_H-d_{th}\right)^2}{16 \mathbb{G}}$.
The last inequality follows from the concentration of the MMD estimate and is true for $n > \frac{64 \mathbb{G}}{(d_H - d_{th})^2}$. The details are provided in Lemma \ref{lem:mmd-dh} in the Appendix.

\begin{align*}
\mathbb{P}\left[\hat{d}_I>d_{t h}\right] &= \mathbb{P}\left[\max _k \max _{\substack{P_{1k}, P_{2k} \\ P_{1k} \cup P_{2k} = P_k \\ P_{1k} \bigcap P_{2k} = \phi}} \min _{\substack{i, j \\ i \in P_{1k} \\ j \in P_{2k}}} \hat{d}(i, j, n)>d_{th}\right] \\
&=\mathbb{P}\left[\bigcup_k \bigcup_{\substack{P_{1k}, P_{2k} \\ P_{1k} \cup P_{2k} = P_k \\ P_{1k} \bigcap P_{2k} = \phi}} \min _{\substack{i, j \\ i \in P_{1k} \\ j \in P_{2k}}} \hat{d}(i, j, n)>d_{th}\right] \\
& \leq \sum_k \sum_{P_{ik}, P_{2k}} \mathbb{P}\left[\min _{\substack{i, j \\ i \in P_{1k} \\ j \in P_{2k}}} \hat{d}(i, j, n)>d_{th}\right] \\
& \text{(using union bound)} \\
& \leqslant \sum_k \sum_{P_{ik}, P_{2k}} \mathbb{P}\left[ \hat{d}(i, j, n)>d_{th}\right] \\
& \text{(Here $i, j$ are chosen such that $d(i,j, n) < d_I$.} \\
& \text{Such a choice exists by the definition of $d_I$.)} \\
& \leqslant K \cdot 2^M \cdot \left(2 \exp \left(-\frac{n\left(d_{th}-d_I\right)^2}{16 \mathbb{G}}\right)\right) \\
& =a_I e^{-b_I n}, 
\end{align*}
where $a_I = 2K 2^M$ and $b_I = \frac{\left(d_{th}-d_{I}\right)^2}{16 \mathbb{G}}$. 
The last inequality follows from the concentration of the MMD estimate and is true for $n > \frac{64 \mathbb{G}}{(d_{th} - d_{I})^2}$. The details are provided in Lemma \ref{lem:mmd-di} in the Appendix.

Combining the bounds above for $\mathbb{P}[\hat{d}_H < d_{th}]$ and $\mathbb{P}[\hat{d}_I > d_{th}]$, we get
\begin{equation*}
\mathbb{P}[\hat{d}_I < \hat{d}_H] \ge (1-a_Ie^{-b_In})(1-a_He^{-b_Hn}),
\end{equation*}
and
\begin{equation*}
P_e \le a_I e^{-b_I n} + a_H e^{-b_H n} \le a_f e^{-b_f n}, 
\end{equation*}
for $n > \max\left(\frac{64 \mathbb{G}}{(d_{H} - d_{th})^2},\frac{64 \mathbb{G}}{(d_{th} - d_{I})^2}\right)$, where $a_f = 2 \max(a_I, a_H)$ and $b_f = \min(b_I,b_H)$.
Thus, we have $P_e \rightarrow 0$ exponentially as $n \rightarrow \infty$.
\end{proof}

\begin{remark}
    Note that $d_{th}$ is chosen such that it satifies $d_I < d_{th} < d_H$. One choice is $d_{th} = (d_I + d_H)/2$. For this choice of $d_{th}$, we get
    \[
      b_f = \frac{(d_H - d_I)^2}{64 \mathbb{G}}.
    \]
    We will use this choice later in the proof of universal consistency of SLINK-SEQ in Theorem \ref{thm:seq-consistency}.
\end{remark}

\begin{remark}
    We mainly discuss SLINK in this paper. We can also extend our $d_I < d_H$ condition to another hierarchical clustering algorithm named complete-linkage based clustering or CLINK. For CLINK, it can be shown that $d_I = d_L$. This means that for CLINK we get back the original $d_L < d_H$ condition proved in \cite{wang2020exponentially,wangthesis}. 
\end{remark}

\section{Proposed Sequential nonparametric clustering algorithm}
\label{sec:seqSLINK}
In this section, we propose a sequential nonparametric clustering algorithm SLINK-SEQ based on SLINK (See Algorithm 1). Estimates of MMD (in equation (\ref{eqn:mmdest})) or KSD (in equation (\ref{eqn:ksdest})) can be used in this algorithm for $\hat{d}(i,j,n)$. To obtain the MMD estimate, at least 2 samples are required. 
The pairwise distances between data sequences, $\{\hat{d}(i,j,n),$ for all $i,j \in \{1,2,...,M\}, i<j\} $, are updated using the new samples in a recursive way to reduce complexity. This sequential update for the MMD $M_b(.)$ is given by:
\begin{align}
M_b(i, j, n)= & \left\{ \left(\frac{n-1}{n}\right)^2 M_b^2(i, j, n-1) \right. \label{mmdupdate}\\
& +\frac{1}{n^2} \sum_{l=1}^n h\left(X_l^{(i)}, X_n^{(i)}, X_l^{(j)}, X_n^{(j)}\right)\hspace{-0.8 in } \nonumber\\
& \left.+\frac{1}{n^2}\sum_{m=1}^{n-1} h\left(X_n^{(i)}, X_l^{(i)}, X_n^{(j)}, X_l^{(j)}\right)\right\}^{\frac{1}{2}}\hspace{-0.8 in } \nonumber
\end{align} 
For KSD, the empirical CDF is updated as:
\[
\hat{F}(n, a) = \frac{n - 1}{n} \hat{F}_X{}_i^{(n-1)}(a)  \frac{1}{n} I(-\infty, a] (X(i)_n), \quad a \in \mathbb{R}.
\]
The stopping rule is proposed to be a threshold on the minimum inter-cluster distance of the clustering output using SLINK at each time. The threshold $T_n$ is chosen to be $C/\sqrt{n}$, where $C > 0$ is a constant. For this choice of threshold, we are able to prove exponential consistency for SLINK-SEQ that uses the MMD estimate later in this section. We also observe that the bias in the MMD estimate is of the order or $1/\sqrt{n}$ in the concentration result for the MMD estimate in \cite[Thm. 7]{JMLR:v13:gretton12a}. Furthermore, in the simulation results section, we compare the performance for different thresholds of the form $C/n^{\alpha}$ with different $\alpha$ and observe that $\alpha = 0.5$ performs the best. 
\vspace{0.5 cm}
\begin{algorithm}
\caption{Proposed SLINK-SEQ}

1: ${\textbf{Input: }}{X^{(1)},...,X^{(M)}}; K$\\
2: ${\textbf{Output: }}$ Clusters ${C_1(N),...,C_K(N)}$\\
3: $\textbf{Initialize: } n \leftarrow 2$, $2$ samples per sequence, $T_2 = 0$\\
4: $\textbf{Calculate distances: } \hat{d}(i,j,n)$ for $i<j$, $i,j\in \{1, 2,\hdots, M\}$\\
5: $\textbf{while } {\Gamma_n < \frac{C}{\sqrt{n}}}$\\
6: \hspace{0.5cm}\textit{\emph{Update clusters} $\{C_k(n)\}_{k=1}^K$ \emph{using SLINK.}}\\
7: \hspace{0.5cm}\textit{\emph{Update test statistic} \\
\hspace{1cm} $\Gamma_n = \min_{k\ne l}\min_{i \in C_k (n), j\in C_l (n)} \hat{d}(i,j,n).$ \\
8: \hspace{0.5cm}\textit{\emph{Continue sampling:}} $n \leftarrow n + 1$ \\ \hspace{1cm} \textit{\emph{and update distances}} \\
9: \textit{\textbf{\emph{end while }} \emph[{Stop sampling} $N = n$\emph]}}
\end{algorithm}


\subsection{Analysis of SLINK-SEQ}

 First, we start with the following assumption.
 \begin{assumption}
  \label{assumption1}
   $d_I < d_H$.
 \end{assumption}
\noindent Given Assumption \ref{assumption1}, we can choose $\delta$ such that   
\[d_I < (1-\delta)^2 d_H,
\] 
i.e., 
\[0 < \delta < 1- \sqrt{\frac{d_I}{d_H}}.\]

Next, we show in Theorem \ref{thm:finite} that the proposed SLINK-SEQ algorithm stops in finite time with probability one.{\footnote{Since one new sample is observed for each data sequence at each time, stopping in finite time is the same as stopping after observing a finite number of samples of each data sequence.}} Here again, we show the analysis only for the SLINK-SEQ algorithm with MMD estimates. Similar results can be obtained for the SLINK-SEQ algorithm with KSD estimates. In the simulation results section, we show results for cases.
\begin{theorem}
The proposed sequential clustering test stops in finite time almost surely, for any set of true clusters $\{P_k : k = 1, \hdots ,K\}$ satisfying Assumption \ref{assumption1}. That is, $P[N < \infty] = 1.$    
\label{thm:finite}
\end{theorem}
\begin{proof}
Let $E_n$ denote the event that clustering output $\{C_1(n), \hdots,C_K(n)\}$ at step $n$ is not equal to the true clusters $\{P_1,\hdots,P_K\}$. In the following, we split $\mathbb{P}[N>n]$ into two terms corresponding to two disjoint events. In the first term, there is a clustering error event after observing $n$ samples, and in the second term there is no clustering error event after observing $n$ samples. The first term can be upper bounded using the FSS upper bound on error probability in Theorem \ref{thm:fssconsistency}. The second term can be bounded by using the convergence properties of the MMD estimates and the fact that the minimum inter-cluster distance is $d_H$.
\begin{align}
\mathbb{P}&[N>n] \nonumber\\
& =\mathbb{P}\left[\{N>n\} \cap E_n\right]+\mathbb{P}\left[\{N>n\} \cap E_n^c\right] \nonumber\\
& \leq \mathbb{P}\left[E_n\right]+\mathbb{P}\left[\{N>n\} \cap E_n^c\right] \nonumber \\
& =\mathbb{P}\left[E_n\right]+\mathbb{P}\left[\left\{\Gamma_{n_1}<T_{n_1}, \forall n_1 \leq n\right\} \cap E_n^c\right] \nonumber\\
& \leq \mathbb{P}\left[E_n\right]+\mathbb{P}\left[\left\{\Gamma_n<T_n\right\} \cap E_n^c\right]\nonumber \\
& \leq \mathbb{P}\left[E_n\right]+\mathbb{P}\left[\left\{\min _{k \neq l} \min _{i \in C_k(n)} \min _{j \in C_l(n)} \hat{d}(i, j, n)<T_n\right\} \cap E_n^c\right] \nonumber\\
& \leq \mathbb{P}\left[E_n\right]+\mathbb{P}\left[\left\{\min _{k \neq l} \min _{i \in P_k} \min _{j \in P_l} \hat{d}(i, j, n)<T_n\right\}\right] \nonumber\\
& \leq \mathbb{P}\left[E_n\right]+\mathbb{P}\left[\left\{\bigcup_{k \neq l} \bigcup_{i \in P_k} \bigcup_{j \in P_l} \hat{d}(i, j, n)<T_n\right\}\right] \nonumber \\
& \leq \mathbb{P}\left[E_n\right]+\sum_{k \neq l} \sum_{i \in P_k} \sum_{j \in P_l} P\left[\hat{d}(i, j, n)<T_n\right]. \label{eqn:stopping}
\end{align} 
The first term in (\ref{eqn:stopping}) is bounded as in Theorem \ref{thm:fssconsistency} to get:
\[
\mathbb{P}\left[E_n\right] \le a_f e^{-b_f n},
\]
for $n > \max\left(\frac{64 \mathbb{G}}{(d_{H} - d_{th})^2},\frac{64 \mathbb{G}}{(d_{th} - d_{I})^2}\right)$.
Therefore, this term goes to zero as $n$ goes to infinity. Now, consider the second term in (\ref{eqn:stopping}).
\begin{align}
\mathbb{P}&\left[ \hat{d}(i, j, n)<T_n\right]\nonumber \\
& = \mathbb{P}\left[ - \hat{d}(i, j, n) > -T_n\right]\nonumber\\
& =\mathbb{P}\left[d\left(p_i, p_j\right)-d(i, j, n)>d\left(p_i, p_j\right)-T_n\right]\nonumber \\
& \leq \mathbb{P}\left[d\left(p_i, p_j\right)-\hat{d}(i, j, n)>d_H-T_n\right]\nonumber \\
& \leq \mathbb{P}\left[\left|d\left(p_i, p_j\right)-\hat{d}(i, j, n)\right|>d_H-T_n\right]\nonumber
\end{align}
Now, we use the concentration result \cite[Thm. 7]{JMLR:v13:gretton12a}
\[
\mathbb{P}\left[\left|d\left(p_i, p_j\right)-\hat{d}(i, j, n)\right|>4 \sqrt{\frac{\mathbb{G}}{n}} +\epsilon \right] \leq 2 \exp\left(\frac{-n \epsilon ^2}{4 \mathbb{G}}\right).
\]
Let $\epsilon = \frac{d_H - T_n}{2}$. Choose $n$ to be large enough such that
$4 \sqrt{\frac{\mathbb{G}}{n}}+ \epsilon < (1 - \frac{\delta}{2})d_H - T_n$. In order to satisfy this inequality, we need:
\begin{align}
  &4 \sqrt{\frac{\mathbb{G}}{n}}+ \frac{d_H - T_n}{2} < \left(1 - \frac{\delta}{2}\right)d_H - T_n  \nonumber \\
   & \text{(or)} ~4 \sqrt{\frac{\mathbb{G}}{n}} < \frac{d_H - T_n}{2} - \frac{\delta}{2} d_H \label{eqn:dh-tndelta}\\
   & \text{(or)} ~8 \sqrt{\frac{\mathbb{G}}{n}} + \frac{C}{\sqrt{n}} < (1 - \delta) d_H \nonumber \\
   & \text{(or)} ~ n > \left(\frac{C + 8 \sqrt{\mathbb{G}}}{(1-\delta)d_H}\right)^2. \nonumber
\end{align}
Therefore, for $n > \left(\frac{C + 8 \sqrt{\mathbb{G}}}{(1-\delta)d_H}\right)^2$, we have  
\[
\mathbb{P}[\hat{d}(i,j,n) \leq T_n] \leq 2 \exp\left(\frac{-n(d_H - T_n)^2}{16 \mathbb{G}}\right).
\]
For the above inequality, we also need $d_H - T_n > 0$ or $n > \frac{C^2}{d_H ^2}$. This is automatically satisfied for $n > \left(\frac{C + 8 \sqrt{\mathbb{G}}}{(1-\delta)d_H}\right)^2$.
From equation (\ref{eqn:dh-tndelta}), we have
\[
d_H - T_n > 8 \sqrt{\frac{\mathbb{G}}{n}} + \delta d_H > \delta d_H.
\]
Therefore, we have 
\[
\mathbb{P}[\hat{d}(i,j,n) \leq T_n] \leq 2 \exp\left(\frac{-n(\delta d_H)^2}{16 \mathbb{G}}\right).
\]
Using the above bound for $\mathbb{P}[\hat{d}(i,j,n) \leq T_n]$, we get a bound for the second term in (\ref{eqn:stopping}) as follows:
\[
\mathbb{P}\left[\{N>n\} \cap E_n^c\right] \leq M^2 \left(2 \exp\left(\frac{-n(\delta d_H)^2}{16 \mathbb{G}}\right)\right),
\]
which goes to zero as $n \rightarrow \infty $. 
Overall, we get the bound:
\begin{equation}
\mathbb{P}[N > n] \le a_f e^{-b_f n} + 2M^2 \exp\left(\frac{-n(\delta d_H)^2}{16 \mathbb{G}}\right) \label{eqn:pNnbound}    
\end{equation}
for\footnote{Note that the $n_M$ defined in Theorem \ref{thm:seq-consistency} satisfies this condition.} $n > \max\left\{\frac{64 \mathbb{G}}{(d_{H} - d_{th})^2},\frac{64 \mathbb{G}}{(d_{th} - d_{I})^2}, \left(\frac{C + 8 \sqrt{\mathbb{G}}}{(1-\delta)d_H}\right)^2\right\}$.
Therefore, we have 
\[
\mathbb{P}[N > n] \rightarrow 0 ~ \text{as} ~ n \rightarrow \infty,
\]
or, equivalently, $\mathbb{P}[N < \infty]  = 1$.
\end{proof}

Next, we analyze the probability of error of SLINK-SEQ. In Theorem \ref{thm:seq-consistency}, we show that SLINK-SEQ is universally consistent.
\begin{theorem}
The proposed sequential clustering tests are universally consistent under any configuration of the true clusters $\{P_k : k = 1,\hdots,K\}.$ That is, $\displaystyle \lim _{C \rightarrow \infty} P_{\max }=0.$
\label{thm:seq-consistency}
\end{theorem}

\begin{proof}
    Define 
    \[
    \tilde{n} = \frac{\max\left(\frac{64 \mathbb{G}}{(d_H - d_0)^2},\frac{64 \mathbb{G}}{(d_0 - d_I)^2}\right)}{(1 - \delta)^2},
    \]
    \[
    n_M = \left(\sqrt{\Tilde{n}} + \frac{C}{(1-\delta)d_H}\right)^2,
    \]
    and
    \[
    C_M = \frac{(1 + \delta)\sqrt{\Tilde{n}}d_I +8 \sqrt{\mathbb{G}}}{(1 - \delta) - \frac{d_I}{(1 - \delta)d_H}}.
    \]
    
     Let $C > C_M.$ We have 
\begin{align}
    \mathbb{P}[E]=&\sum_{n=2}^{\infty} \mathbb{P}\left[N=n, E_n\right]\nonumber\\
&=\sum_{n=2}^{n_M} \mathbb{P}\left[N=n, E_n\right]+\sum_{n>n_M}^{\infty} \mathbb{P}\left[N=n, E_n\right] \label{eqn:seq-ucbound}
\end{align}
In (\ref{eqn:seq-ucbound}), the first term corresponds to the case where the stopping time is less than or equal to $n_M$ and the second term corresponds to the case where the stopping time is greater than $n_M$. For the second term, we bound the error probability using the FSS error probability bound for large enough $n$.

Consider the second term of (\ref{eqn:seq-ucbound}). This term can be bounded as follows.
\begin{align}
&\sum_{n>n_M}^{\infty} \mathbb{P}\left[N=n, E_n\right] \le \sum_{n>n_M}^{\infty} \mathbb{P}\left[E_n\right] \nonumber \\
& = \sum_{n>n_M}^{\infty} a_f e^{-b_f n} \nonumber \\
& = \frac{a_f}{(1 - e^{-b_f})} e^{-b_f n_M} \nonumber \\
& \le \frac{a_f}{(1 - e^{-b_f})} e^{-b_f \frac{C^2}{(1-\delta)^2 d_H^2}}, \nonumber
\end{align}
where the last inequality is obtained using $n_M > \frac{C}{(1-\delta) d_H}$. Therefore, we have that the second term in (\ref{eqn:seq-ucbound}) goes to 0 as $C \rightarrow \infty$. 

Now, consider the first term in (\ref{eqn:seq-ucbound}). We show that this term can be bounded for large enough $C$, and using the definition of $d_I$. The choice of $C_M$ and $n_M$ are important to achieve this goal.
\begin{align}
& \sum_{n=2}^{n_M} \mathbb{P}\left[N=n, E_n\right] \le \sum_{n=2}^{n_M} \mathbb{P}\left[\Gamma_n>T_n ,E_n\right]\nonumber \\
&=\sum_{n=1}^{n_M} \mathbb{P}\left[\left\{\min _{k \neq l}\min _{i \in C_k(n)} \min _{j \in C_l(n)} \hat{d}(i, j, n)>T_n , \forall k, l\right\} \bigcap E_n\right] \nonumber\\
&\leq \sum_{n=2}^{n_M} \mathbb{P}\left[ \hat{d}(i, j, n)>T_n \bigcap E_n\right] \label{eqn:use-di-defn} \\
&\leq \sum_{n=2}^{n_M} \mathbb{P}\left[ \hat{d}(i, j, n)>T_n \right], \nonumber
\end{align}   
where, in (\ref{eqn:use-di-defn}), $i$, $j$ are chosen such that the true distance between the sequences $d(p_i,p_j) < d_I$, both sequences $i$, $j$ are from the same true cluster but clustered by the algorithm in different clusters. Such a choice is always possible due to the definition of $d_I$ and the fact that $E_n$ is true. Since $E_n$ is true, there must be at least one cluster that is not correctly identified, i.e., it is partitioned into at least 2 sub-clusters. Now, by the definition of $d_I$, for each sub-cluster, there must be at least one other sub-cluster such that two sequences from these two sub-clusters are within $d_I$ of each other. 
Continuing, we have
\begin{align}
    &\mathbb{P}[\hat{d}(i,j,n) >T_n] \nonumber \\
    & = \mathbb{P}[\hat{d}(i,j,n) - d(p_i,p_j) > T_n - d(p_i,p_j)] \nonumber\\
    & \leq \mathbb{P}[\hat{d}(i,j,n) - d(p_i,p_j) > T_n - d_I] \nonumber \\
    & \leq \mathbb{P}[|\hat{d}(i,j,n) - d(p_i,p_j)| > T_n - d_I]. 
\end{align}
Let $\epsilon = \frac{T_n - d_I}{2}$. We would like to now apply the concentration result to get 
\begin{align}
&\mathbb{P}[|\hat{d}(i,j,n) - d(p_i,p_j)| > T_n - d_I] \nonumber\\
&\le 2 \exp\left(\frac{-n(T_n - d_I)^2}{16\mathbb{G}}\right) \label{eqn:concstep1}\\
& \le 2 \exp\left(\frac{-n(\delta T_n)^2}{16\mathbb{G}}\right) \label{eq:concstep2}\\
& = 2 \exp\left(\frac{-\delta^2 C^2}{16\mathbb{G}}\right).
\end{align}
In order to get (\ref{eqn:concstep1}) and (\ref{eq:concstep2}), we need the following condition:
\[
4 \sqrt{\frac{\mathbb{G}}{n}} + \epsilon < (1-\frac{\delta}{2})T_n - d_I < T_n - d_I\]
for $n \le n_M$. We can rewrite this condition as follows.
\begin{align*}
   & 4 \sqrt{\frac{\mathbb{G}}{n}} + \frac{T_n - d_I}{2} < T_n - d_I -\frac{\delta}{2}T_n\\
   &\text{(or)} ~ 4 \sqrt{\frac{\mathbb{G}}{n}} < \frac{1-\delta}{2} T_n - \frac{d_I}{2}\\
   & \text{(or)} ~ 8 \sqrt{\frac{\mathbb{G}}{n}} < (1 - \delta) \frac{C}{\sqrt{n}} - d_I \\
   & \text{(or)} ~ C > \frac{d_I \sqrt{n} + 8 \sqrt{\mathbb{G}}}{(1- \delta)}
\end{align*}
The above condition can be satisfied for all $n \le n_M$, if it is satisfied for $n = n_M$, i.e., 
\begin{align}
&   C > \frac{d_I \sqrt{n_M} + 8 \sqrt{\mathbb{G}}}{(1- \delta)} \nonumber \\ 
& \text{(or)} ~ (1 - \delta) C > d_I \left( \sqrt{\tilde{n}} + \frac{C}{(1-\delta)d_H}\right) + 8 \sqrt{\mathbb{G}} \nonumber \\
& \text{(or)} ~ C > \frac{d_I \sqrt{\Tilde{n}} + 8 \sqrt{\mathbb{G}}}{(1 - \delta ) - \frac{d_I}{(1 - \delta)d_H}} \label{eqn:CM}
\end{align}
   
Note that the denominator $(1 - \delta ) - \frac{d_I}{(1 - \delta)d_H}> 0$. The condition in (\ref{eqn:CM}) is satisfied for $C > C_M$. Therefore, we have 
\begin{align}
    \mathbb{P}[\hat{d}(i,j,n)>T_n]& \leq  2 \exp\left(-\frac{\delta ^2 C^2}{16 \mathbb{G}}\right) 
\end{align}
and the first term in (\ref{eqn:seq-ucbound}) can be bounded as
\begin{equation}
    \sum_{n=2}^{n_M} \mathbb{P}\left[ \hat{d}(i, j, n)>T_n \right] \le 2 n_M  \exp\left(-\frac{\delta ^2 C^2}{16 \mathbb{G}}\right)
\end{equation}
for $C > C_M$. Note that $n_M$ is only quadratic with $C$. Therefore, as $C \rightarrow \infty$, this term also goes to 0. 

Since (\ref{eqn:seq-ucbound}) goes to 0 for any configuration of 
true clusters, we have universal consistency, i.e., $\displaystyle \lim _{C \rightarrow \infty} P_{\max }=0.$ 
\end{proof}

Theorem \ref{thm:seq-consistency} shows universal consistency of  SLINK-SEQ. We can observe that both the terms in the bound on probability of error (in (\ref{eqn:seq-ucbound}))
\begin{equation}
\mathbb{P}[E] \le \frac{a_f}{(1 - e^{-b_f})} e^{-b_f \frac{C^2}{(1-\delta)^2 d_H^2}} + 2 n_M  \exp\left(-\frac{\delta ^2 C^2}{16 \mathbb{G}}\right)
    \label{eqn:pebound}
\end{equation}
decrease exponentially with $C^2$ for large $C$. For exponential consistency, we need to show that the sequence of tests with increasing $C$ has error probability that decreases exponentially with expected stopping time, i.e., 
\[
\mathbb{E}[N] \le \frac{-\log P_{max}}{\alpha} (1 + o(1)),
\]
where $\alpha$ is strictly positive. We proceed to show this using steps similar to that in \cite{sreenivasan2023nonparametric,doi:10.1080/07474946.2017.1360086} in the following two theorems. We skip some of the details that are similar, and retain the main arguments for completeness. Theorem \ref{thm:NC2} follows from: (1) the choice of threshold $C/\sqrt{n}$, and (2) uniform integrability of $N/C^2$. Theorem \ref{thm:exp-cons-seq} relies on the exponential bound derived in (\ref{eqn:pebound}) and Theorem \ref{thm:NC2}.

\begin{theorem}
The stopping time $N$ of SLINK-SEQ satisfies: $\lim_{C \rightarrow \infty} \mathbb{E}[\left| \frac{N}{C^2} - \frac{1}{d_H^2} \right|] = 0.$
    \label{thm:NC2}    
\end{theorem}
\begin{proof}
    The algorithm stops when the minimum inter-cluster distance of the clustering output $\Gamma_n$ crosses the threshold $T_n = \frac{C}{\sqrt{n}}$. Therefore, with probability 1, we have 
    \[
    \frac{1}{\Gamma_N^2} \le \frac{N}{C^2} \le \frac{1}{\Gamma_{N-1}^2} + \frac{1}{C^2}.
    \]
    Since the kernel is bounded, the MMD estimate $\hat{d}(i,j,n)$ is also bounded, say by $B_d$. Therefore, for $n < C^2/{B_d^2}$, $\mathbb{P}[N < n] = 0$. Thus, $N \rightarrow \infty$ as $C \rightarrow \infty$. From \cite[Thm. 7]{JMLR:v13:gretton12a} and Borel-Cantelli Lemma, we have $\Gamma_n^2 \rightarrow d_H^2$ almost surely as $n \rightarrow \infty$. Therefore, we have 
    \[
    \frac{N}{C^2} \rightarrow \frac{1}{d_H^2} ~~\text{a.s.}
    \]
    as $C \rightarrow \infty$. In order to complete the proof, we need to show that $N/C^2$ is uniformly integrable, or, equivalently 
    \[
    \lim_{\nu \rightarrow \infty} \sup_{C \ge C_M} \mathbb{E}\left[ \frac{N}{C^2} \mathbb{I}\left( \frac{N}{C^2} \ge \nu\right)\right] = 0.
    \]
    Following the simplifications in \cite[Thm. 3]{sreenivasan2023nonparametric} for $\mathbb{E}\left[ \frac{N}{C^2} \mathbb{I}\left( \frac{N}{C^2} \ge \nu\right)\right]$, and using the bound for $\mathbb{P}[N > n]$ from (\ref{eqn:pNnbound}), we can show that:
    \begin{align*}
    &\sup_{C \ge C_M} \mathbb{E}\left[ \frac{N}{C^2} \mathbb{I}\left( \frac{N}{C^2} \ge \nu\right)\right] \le \frac{1}{C_M^2} \left[ \frac{a_f e^{-b_f \lfloor{\nu C_M^2}\rfloor}}{1 - e^{-b_f}} \right.\\
    & \left. + \frac{2M^2 e^{-\frac{\delta^2 d_H^2}{16 \mathbb{G}} \lfloor \nu C_M^2 \rfloor }}{ 1 - e^{-\frac{\delta^2 d_H^2}{16 \mathbb{G}}}}\right] + \nu \left[ a_f e^{-b_f(\lfloor{\nu C_M^2}\rfloor - 1)}\right. \\
    & \left. + 2M^2 e^{-\frac{\delta^2 d_H^2}{16 \mathbb{G}} (\lfloor{\nu C_M^2}\rfloor - 1)}\right],    
    \end{align*}
    and this goes to zero as $\nu \rightarrow \infty$.
\end{proof}

\begin{theorem}
    SLINK-SEQ is exponentially consistent, i.e., 
    \[
    \mathbb{E}[N] \le \frac{-\log P_{max}}{\alpha} (1 + o(1)),
    \]
    with $\alpha > 0$.
    \label{thm:exp-cons-seq}
\end{theorem}
\begin{proof}
    We know that $N/C^2$ converges to $1/d_H^2$ as $C \rightarrow \infty$ as in Theorem \ref{thm:NC2}. From (\ref{eqn:pebound}), we know that $\frac{-\log P_{max}}{C^2}$ is upper bounded by a constant given by 
    \[
    \alpha_1 = \min\left\{ \frac{\delta^2}{16\mathbb{G}}, \frac{b_f}{(1 - \delta)^2 d_H^2}\right\}
    \]
    for large $C$. Combining these two results, we can show exponential consistency with $\alpha = \alpha_1 d_H^2$.
\end{proof}

\subsection{Complexity of the SLINK-SEQ}
We now discuss the computational complexity of the SLINK-SEQ algorithm. At each time $n$, we do the following computations.
\begin{itemize}
    \item Updating the pairwise distances between the data sequences: There are $M(M-1)/2$ distances to be updated. {The MMD can be updated using the sequential update in $O(n)$ kernel computations. The kernel can computed in $O(n_{\text{dim}})$ computations, where $n_{\text{dim}}$ is the dimension of each sample of a data sequence. Therefore, the overall complexity of this step is of the order $O(dn_{\text{dim}}M^2)$.}
    \item Updating the clusters using SLINK: SLINK starts from $M$ clusters and reduces the number of clusters one at a time to $K$ clusters. Therefore, there are $M-K$ steps. In each step of SLINK, the distance of the merged cluster with the other clusters needs to be updated. This would require $O(M-i)$ computations at step $i$. The overall complexity of this step is $O(M^2)$.
\end{itemize}
Overall, the complexity grows at most quadratically in the number of data sequences $M$ {and linearly in the number of dimensions $n_{\text{dim}}$ of each sample in a data sequence}. Further simplification could be studied in future work.

{The cost of sequential MMD update in (\ref{mmdupdate}) grows as more data is observed, as it requires $O(n)$ complexity. Some simplifications of this by using only a limited window of prior samples in the MMD update could be considered to reduce this complexity. However, further investigation is required to identify the right window size. This issue and other related issues on the scalability of clustering algorithms have been studied in \cite{9440980scale}.
}

\section{Simulation Results}
\label{sec:simulations}
We now study the performance of SLINK and SLINK-SEQ using simulations.\footnote{The code used to generate the results in this paper are available at https://github.com/bhupenderee22s006/TSP\_codes.git } For the FSS setting, we illustrate that SLINK is exponentially consistent under the condition $d_I < d_H$. We also compare SLINK with $k$-medoids clustering (KMed) in \cite{8651294}. Then, we compare SLINK with SLINK-SEQ and show how sequential clustering can provide improved performance. In the simulations, we run the clustering algorithms for multiple realizations of the data sequences and compare the final clustering output to the true clusters. An error is declared if the clustering output does not exactly match the true clusrers. We estimate $P_e$ by taking the ratio of the number of errors to the number of realizations simulated. In the case of sequential algorithms, we also determine the number of samples at which the algorithm makes a decision on the clusters for each realization. $\mathbb{E}[N]$ is estimated as the average number of samples taken over all the realizations.

\subsection{Simulation settings}
First, we consider three examples to illustrate the difference between the distance assumptions between the clusters. Examples 1 and 2 have $d_L > d_H$ but $d_I < d_H$. Example 3 satisfies $d_L < d_H$ and has been used in \cite{8651294}. 
 In each example, the i.i.d. data sequences are generated according Gaussian distributions with unit variance and different means. The three examples are: (1) 2 clusters, means of the sequences in cluster 1 and cluster 2 are $\{0.4, 0.55, 0.7, 0.85, 1, 1.15, 1.3, 1.45, 1.6\}$ and $\{1.85, 2, 2.15\}$, (2) 2 clusters, means of the sequences in cluster 1 and cluster 2 are $\{0.7, 0.85, 1, 1.15, 1.3\}$ and $\{1.7, 1.85, 2, 2.15, 2.3\}$, and (3) 5 clusters, 5 sequences per cluster, mean of each sequence in cluster $k$ is $k-1$ for $k = 1, 2, 3, 4, 5$.

For the above examples, $d_I$ is equal to the maximum gap between two neighbouring distributions within a cluster, i.e, for each distribution we need only the left and right neighbours to be closer than the distributions in the other clusters. Therefore, we can have big clusters that are separated by gaps that are small (as in Example 1 and 2). In fact, for these examples, clustering is achieved by identifying the largest gap between means. Table \ref{d_table} shows the MMD and KSD distances $d_L$, $d_H$, and $d_I$ for each of the three examples. The distances are estimated using 10000 samples from each distribution.\footnote{Analytical evaluation of MMD is not straighforward and has been studied in \cite{mmdeval10.5555/2999134.2999136} The MMD between distributions $p$ and $q$ can be calculated in terms of expected kernels for samples from $(p,  p)$, $(q, q)$ and $(p, q)$. In Table 1 of \cite{mmdeval10.5555/2999134.2999136}, the analytical form of the expected kernel is provided some cases of $p$, $q$ and kernel function choices including the special case where $p$ and $q$ are Gaussian and the kernel is also Gaussian. We have checked that the values we report in Table II of our paper match with the distance evaluated using the expressions in \cite{mmdeval10.5555/2999134.2999136}.} Clearly, we have $d_I < d_H$ in all three examples, while $d_L < d_H$ is true only in example 3.

\begin{figure}[!ht]
    \centering
    \includegraphics[width=0.5\textwidth]{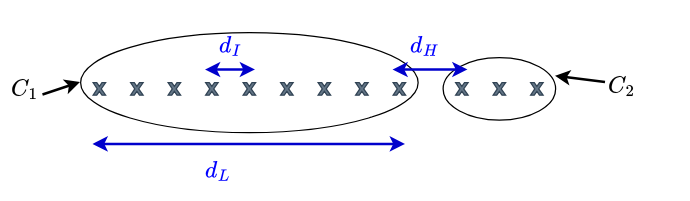}
    \caption{Example 1: One big cluster and one small cluster, the means of the data sequences are shown, $d_L=0.49401$, $d_I = 0.06238$, $d_H = 0.11152$ using MMD. Here, $d_L > d_H$ but $d_I < d_H$. Note that the distances $d_I$, $d_L$ and $d_H$ indicated are actually the distances between the distributions and not Euclidean distance.}
    \label{fig:image1}
\end{figure}
\begin{figure}[!ht]
    \centering
    \includegraphics[width=0.5\textwidth]{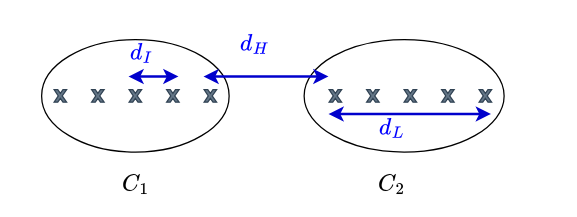}
    \caption{Example 2: Two equal sized clusters, the means of the data sequences are shown, $d_L=0.26219$, $d_I = 0.06238$, $d_H = 0.1665$ using MMD. Here, $d_L > d_H$ but $d_I < d_H$.}
\end{figure}
\begin{figure}[!ht]
    \centering
    \includegraphics[width=0.5\textwidth]{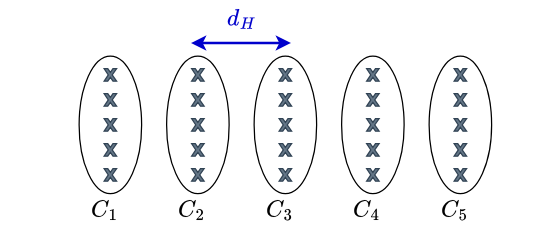}
    \caption{Example 3: 5 clusters, the means of the data sequences are shown, $d_L=0$, $d_I = 0$, $d_H = 0.41289$ using MMD. Here, $d_I = d_L < d_H$}
    \label{fig:image2}
\end{figure}

\begin{table}[!hb]
\centering
\renewcommand{\arraystretch}{1.5} 
\caption{Distances $d_L$, $d_H$ and $d_I$ for Gaussian Examples 1-3}
\begin{tabular}{|c|c|c|c|c|c|c|}
\hline
         & \multicolumn{3}{|c|}{\textbf{MMD}} & \multicolumn{3}{|c|}{\textbf{KSD}} \\
        \hline
            & \textbf{Eg. 1} &  \textbf{Eg. 2} & \textbf{Eg. 3} & \textbf{Eg. 1} & \textbf{Eg. 2}  & \textbf{Eg. 3} \\
        \hline
        $d_L$ & 0.49401 & 0.26219 & 0  & 0.444 & 0.2362 & 0 \\
        \hline
        $d_H$ & 0.11152 &  0.1665 & 0.41289 & 0.0995 &0.1668 & 0.3789 \\
        \hline
        $d_I$ & 0.06238 &  0.06238 & 0 & 0.0541 &0.0541  & 0 \\
        \hline
\end{tabular}
\label{d_table}
\end{table}

\begin{table}[!hb]
\centering
\renewcommand{\arraystretch}{1.5} 
\caption{Distances $d_L$, $d_H$ and $d_I$ for GMM Examples 4-5}
\begin{tabular}{|c|c|c|c|c|c|c|}
\hline
         & \multicolumn{2}{|c|}{\textbf{MMD}} \\
        \hline
            & \textbf{Eg. 4} &  \textbf{Eg. 5}  \\
        \hline
        $d_L$ & 0.41258 & 0.41258 \\
        \hline
        $d_H$ & 0.25897 &  0.35536 \\
        \hline
        $d_I$ & 0.24583 &  0.24583\\ 
        \hline
\end{tabular}
\label{d_table2}
\end{table}
\vspace{1 cm}

Next, we consider two more examples with non-Gaussian data sequences. There are 2 clusters with 3 data sequences each. Here, each data sequence is generated in an i.i.d. manner according to a Gaussian Mixture Model (GMM). In particular, each sequence is a mixture of 2 Gaussian distributions: $N(m_1,1)$ with probability $0.7$, and $N(m_2,1)$ with probability 0.3. For Example 4, the $(m_1, m_2)$ values for the sequences in the two clusters are as follows: \{$(-0.5, 0), (0, 0.5), (0.5, 1)$\}, and \{$(1.2, 1.7), (1.7, 2.2), (2.2, 2.7)$\}. 
For Example 5, the $(m_1, m_2)$ values for the sequences in the two clusters are as follows: \{$(-0.5, 0), (0, 0.5), (0.5, 1)$\}, and \{$(1.35, 1.85), (1.85, 2.35), (2.35, 2.85)$\}. It can be seen that the clusters are closer in Example 4 compared to Example 5. 
The distances $d_L$, $d_I$, and $d_H$ for these two examples are given in Table \ref{d_table2}. In both cases, $d_L > d_H$, but $d_I < d_H$. In particular, $d_I$ is only slightly lesser than $d_H$ in Example 4.

{Finally, we consider two real-world data set examples, one using Modified National Institute of Standards and Technology (MNIST) data \cite{mnist726791}, and the other using MovieLens data \cite{cantador2011second,Kota_Karthik_Tan_2023}.} The MNIST data set contains about 6000 images each of handwritten numerals from 0 to 9 of size $28 \times 28$. Using this data set, we create five data sequences of 1000 samples for each numeral from 0 to 9. Thus, we have 50 data sequences forming 10 clusters. Here, each sample in a data sequence has $28 \times 28 = 784$ dimensions For this multi-dimensional sample case, we use the kernel function: 
\[
k({\mathbf{x}}, {\mathbf{y}}) = e^{-||{\mathbf{x}} - {\mathbf{y}}||^2/2}.
\]
We also created another lower dimensional data set that reduces the 784 dimensions to 10 dimensions using principal component analysis. {The MovieLens data set consists of movie ratings from different viewers of movies from different genres. We take data of five genres and classify then into 3 clusters, i.e., 5 sequences and 3 clusters. The average ratings of the 5 genres: Horror, Adventure, Thriller, Romance and Crime are 3.2034, 3.4010, 3.4318, 3.4666 and 3.6014 respectively, corresponding to 3 clusters \{Crime\}, \{Adventure, Thriller, Romance\} and \{Horror\}.}

\begin{figure}[!ht]
\centering
\begin{tikzpicture}[scale=1]
\begin{axis}
[xlabel = {\large{$n$}}, ylabel = {\large{$\text{ln}(P_e)$}}, xmin =380, xmax=3020,ymin=-3.1, ymax=0.20,
xtick={500,1000,1500,2000,2500,3000},
xticklabels={500,1000,1500,2000,2500,3000},
grid=major,
ytick={-3.0,-2.5,-2.0,-1.5,-1.0,-0.50,0.00,0.50},
yticklabels={-3.0,-2.5,-2.0,-1.5,-1.0,-0.50,0.00,0.50},
legend style={at={(0,0)},anchor=south west,nodes={scale=0.75}}]
\addplot[mark = *,blue,thick, mark options={fill=blue}] coordinates{
(500,-0.58778)
(1000,-0.88376)
(1500,-1.25276)
(2000,-1.68268)
(2500,-2.08318)
(3000,-2.43711)
(3500,-2.9704)
};
\addplot[mark = *,red,thick, mark options={fill=red}] coordinates{
(500,-0.75612)
(1000,-1.05779)
(1500,-1.50851)
(2000,-2.00687)
(2500,-2.38047)
(3000,-2.94443)
};

\addplot[mark = square,teal,thick] coordinates{
(500,-0.06)
(1000,0)
(1500,0)
(2000,0)
(2500,0)
};
\legend{{\large{SLINK, KSD}},\large{SLINK, MMD}, \large{KMed, KSD}}
\end{axis}
\end{tikzpicture}
\caption{FSS performance comparison of SLINK (using MMD or KSD as distance) and KMed for Example 1. Using MMD, we have  $d_L=0.49401$, $d_I = 0.06238$, $d_H = 0.11152$. KMed fails to detect the correct clustering in this case.}
\label{fig:bigsmall}
\end{figure}
\begin{figure}[!ht]
\centering
\begin{tikzpicture}[scale=1]
\begin{axis}
[xlabel = {\large{$n$}}, ylabel = {\large{$\text{ln}(P_e)$}}, xmin =32, xmax=255,ymin=-2.7, ymax=-0.3,
xtick={40,60,100,150,200,250},
xticklabels={30,60,100,150,200,250},
grid=major,
ytick={-4.5,-4.0,-3.5,-3.0,-2.5,-2,-1.5,-1,-0.5,0},
yticklabels={-4.5,-4.0,-3.5,-3.0,-2.5,-2,-1.5,-1,-0.5,0},
legend style={at={(0,0)},anchor=south west ,nodes={scale=0.7}}]
\addplot[mark = *,blue,thick, mark options={fill=blue}] coordinates{
(50, -0.49286529838899795)
(100,-0.8475835338643642)
(150, -1.0483710722313628)
(200, -1.3648153294427663)
(250, -1.6157181503912508)
};
\addplot[mark = *,red,thick, mark options={fill=red}] coordinates{
(50, -0.4934759854308776)
(75, -0.6921466802263617)
(100, -0.870037347356693)
(125, -0.9820784724121581)
(150, -1.1474024528375417)
(175, -1.324551927284321)
(200, -1.4590800308455383)
(225, -1.5896432851059208)
(250, -1.6933193964148026)
};

\addplot[mark =square,teal,thick] coordinates{
(50,-0.90825856)
(100,-1.3912819 )
(150,-1.78808606 )
(200,-2.1344031 )
(250,-2.36292732)
};

\addplot[mark = square,magenta,thick] coordinates{
(50,-0.97229292)
(100,-1.49245456)
(150,-2.0087508)
(200,-2.38010154)
(250,-2.63490587)
};

\legend{{\large{SLINK, KSD}}, \large{SLINK, MMD}, \large{KMed, KSD},\large{KMed, MMD}  }
\end{axis}
\end{tikzpicture}
\caption{FSS performance comparison of SLINK and KMed for Example 2. Using MMD, we have  $d_L=0.26219$, $d_I = 0.06238$, $d_H = 0.1665$.}
\label{fig:eg2}
\end{figure}
\begin{figure}[!ht]
\centering
\begin{tikzpicture}[scale=1]
\begin{axis}
[xlabel = {\large{$n$}}, ylabel = {\large{$\text{ln}(P_e)$}}, xmin =28, xmax=72,ymin=-4.75, ymax=-0.3,
xtick={30,35,40,45,50,55,60,65,70},
xticklabels={30,35,40,45,50,55,60,65,70},
grid=major,
ytick={-4.5,-4.0,-3.5,-3.0,-2.5,-2,-1.5,-1,-0.5},
yticklabels={-4.5,-4.0,-3.5,-3.0,-2.5,-2,-1.5,-1,-0.5},
legend style={at={(0,0)},anchor=south west ,nodes={scale=0.7}}]
\addplot[mark = *,blue,thick, mark options={fill=blue}] coordinates{
(30, -0.41871033485818493)
(40, -1.252762968495368)
(50, -2.0068708488450007)
(60, -2.9927277645336923)
(70, -3.764682417529437)
};

\addplot[mark = *,red,thick, mark options={fill=red}] coordinates{
(30, -0.4382549309311553)
(40, -1.3558351536351823)
(50, -2.1482677326096886)
(60, -2.96424160646262)
(70, -4.070222620101691)
};

\addplot[mark =square,teal,thick] coordinates{
(30,-0.9242589)
(40,-1.57691472)
(50,-2.49979526)
(60,-3.37485341)
 (70,-4.35963494)
};

\addplot[mark = square,magenta,thick] coordinates{
(30, -0.95320109)
(40,-1.82034693)
(50,-2.73332801)
(60,-3.72641606)
(70,-4.69840549)
};

\addplot[mark = square,black,thick] coordinates{
(30, -0.8329091)
(40,-1.591782977)
(50,-2.43031)
(60,-3.17178)
(70,-3.87328)
};

\addplot[mark = *,black,thick, mark options={fill=blue}] coordinates{
(30,-0.9062404)
(40,-1.6534548)
(50,-2.48615587)
(60,-3.23218684)
(70,-4.04637906)
};

\legend{{\large{SLINK, KSD}},\large{SLINK, MMD}, \large{KMed, KSD},\large{KMed, MMD}, \large{CLINK, KSD},\large{CLINK, MMD}   }
\end{axis}
\end{tikzpicture}
\caption{FSS performance comparison of SLINK, KMed and CLINK for Example 3. Using MMD, we have  $d_L=0$, $d_I = 0$, $d_H = 0.41289$.}
\label{fig:eg3}
\end{figure}
\subsection{FSS performance}
Fig. \ref{fig:bigsmall} shows $\text{ln}(P_e)$ versus the sample size $n$ for Example 1. The performance of SLINK is shown for both MMD (with kernel $k(x,y) = e^{-(x-y)^2/2}$) and KSD as the distance measures. It can be observed that SLINK is exponentially consistent, i.e., the $\text{ln}(P_e)$ vs. $n$ plot is linearly decreasing, as expected. For this example, KMed is unable to find the true clusters in the FSS setting for any $n$. Although only the KSD case is shown for KMed, KMed with MMD also does not work. This is mainly due to the fact that one cluster is big while the other is small, and some distributions in the big cluster are closer to the medoid of the small cluster. However, SLINK can work even in this setting as $d_I < d_H$ is satisfied as long as the left and right neighbours for each distribution (in terms of the means of the Gaussian distributions) within the same cluster are closer than the distributions from the other clusters. {In Table \ref{d_table3}, we compare the observed slope in the simulation in Fig. \ref{fig:bigsmall} with the slope of the upper bound in Remark 1. We observe that the simulated performance slope is better than the upper bound by a factor about 25 in this case. The upper bound was mainly derived to show exponential consistency, and provides only a loose bound on the actual slope.}

Figs. \ref{fig:eg2}  and \ref{fig:eg3} show $\text{ln}(P_e)$ versus the sample size $n$ for Examples 2 and 3.  Both SLINK and KMed are exponentially consistent in the FSS setting for these examples. Even though $d_L > d_H$, KMed works for Example 2 since the clusters are of equal size and, therefore, all the distributions in each cluster are closer to the medoid of the correct cluster. For these two examples, KMed is able to perform better than SLINK, i.e., it achieves the same error probability at lower $n$. In particular, for example 2, KMed with MMD estimates can achieve $\text{ln}(P_e) = -1.5$ with 100 samples, while SLINK with MMD estimates requires around 200 samples. However, SLINK can work for all problems with $d_I < d_H$, while KMed cannot. Fig. \ref{fig:eg3} also shows the performance of CLINK using both KSD and MMD estimates. It can be observed that for this example, CLINK is also exponentially consistent. It performs better than SLINK for lower $n$. However, CLINK does not work for Examples 1 and 2 since $d_L > d_H$ in these two examples. {The comparison of the estimated slope from the simulation with the upper bound in Remark 1 is shown for these examples also in Table \ref{d_table3}.}

\begin{table}[!ht]
\centering
\renewcommand{\arraystretch}{1.5} 
\caption{{Comparison of slope: Simulation vs. Analysis}}
\begin{tabular}{|c|c|c|c|c|c|c|}
\hline
         & \multicolumn{2}{|c|}{\textbf{Magnitude of slope of $\ln{P_e}$ vs. $n$}} \\
        \hline
            & $b_f = \frac{(d_H - d_I)^2}{64 \mathbb{G}}$ in bound &  Estimated from Figs. 5-7  \\
        \hline
        Example 1 & $0.3773 \times 10^{-4}$ & $9.572 \times 10^{-4}$ \\
        \hline
        Example 2 & $0.1694 \times 10^{-3}$ & $5.4885 \times 10^{-3}$ \\
        \hline
        Example 3 & $0.2664 \times 10^{-2}$ & $9.084 \times 10^{-2}$\\ 
        \hline
\end{tabular}
\label{d_table3}
\end{table}

\subsection{Sequential clustering performance}
Figs. \ref{fig:seqfss2} shows $\text{ln}(P_e)$ versus $E[N]$ for SLINK-SEQ and SLINK FSS using MMD distance measure for Examples 1 and 2. As expected, we observe that the proposed sequential clustering algorithm requires fewer expected number of samples than the FSS test for the same $P_e$. Similar results are shown for Example 3 in Fig.  \ref{fig:seqfss3}, respectively. Here, we show only the performance with MMD. The performance with KSD shows a similar trend.
\begin{figure}[!ht]
\centering
\begin{tikzpicture}[scale=1]
\begin{axis}
[xlabel = {\large{$\mathbb{E}[N]$}}, ylabel = {\large{$\text{ln}(P_e)$}}, xmin =0, xmax=3300,ymin=-5.2, ymax=-0.5,
xtick={0,500,1000,1500,2000,2500,3000},
xticklabels={0,500,1000,1500,2000,2500,3000},
grid=major,
ytick={-5,-4.5,-4,-3.5,-3,-2.5,-2,-1.5,-1,-0.5,0},
yticklabels={-5,-4.5,-4,-3.5,-3,-2.5,-2,-1.5,-1,-0.5,0},
legend style={at={(1,0)},anchor=south east,nodes={scale=0.7}}]

\addplot[mark = *,red,thick, mark options={fill=red}] coordinates{
(500,-0.75612)
(1000,-1.05779)
(1500,-1.50851)
(2000,-2.00687)
(2500,-2.38047)
(3000,-2.94443)
};

\addplot[mark = triangle,black,thick, mark options={fill=black}] coordinates{
(504.8525,-0.73668)
(821.3233,-1.0767)
(1159.9064,-1.47063)
(1564.7505,-1.93484)
(1978.9848,-2.46631)
(2440.0495,-3.09145)
(2930.441,-3.76930)
};
\addplot[mark = square,red,thick, mark options={fill=red}] coordinates{
(100,-0.916290731874155)
(200,-1.4562867329399256)
(300,-1.9154509415706047)
(400,-2.4773783833672085)
(500,-3.112181086197238)
(600,  -3.6179203651734424)
};

\addplot[mark = square,black,thick, mark options={fill=black}] coordinates{
(64.6331,-0.68200943)
(140.0953,-1.3208812225)
(232.0438,-2.1136189934814222)
(335.63,-3.221)
(455.657,-4.1799)
(555.289,-5.14)
};

\legend{{\large{SLINK, FSS, MMD, Example 1}}, \large{SLINK-SEQ, MMD, Example 1}, {\large{SLINK, FSS, MMD, Example 2}}, \large{SLINK-SEQ, MMD, Example 2}}
\end{axis}
\end{tikzpicture}
\caption{Performance improvement using SLINK-SEQ compared to SLINK FSS for Examples 1 and 2, using MMD as distance. Since Example 2 is easier than Example 1, fewer samples are required for Example 2 to achieve the same performance.}
\label{fig:seqfss2}
\end{figure}
\begin{figure}[!ht]
\centering
\begin{tikzpicture}[scale=1]
\begin{axis}
[xlabel = {\large{$\mathbb{E}[N]$}}, ylabel = {\large{$\text{ln}(P_e)$}}, xmin =29, xmax=71,ymin=-6.50, ymax=-0.4,
xtick={30,35,40,45,50,55,60,65,70},
xticklabels={30,35,40,45,50,55,60,65,70},
grid=major,
ytick={-6.50,-6,-5.5,-5,-4.5,-4,-3.5,-3,-2.5,-2,-1.5,-1,-0.5},
yticklabels={-6.50,-6,-5.5,-5,-4.5,-4,-3.5,-3,-2.5,-2,-1.5,-1,-0.5},
legend style={at={(0,0)},anchor=south west,nodes={scale=0.7}}]

\addplot[mark = *,red,thick, mark options={fill=red}] coordinates{
(30, -0.4382549309311553)
(40, -1.3558351536351823)
(50, -2.1482677326096886)
(60, -2.96424160646262)
(70, -4.070222620101691)
};
\addplot[mark = triangle,black,thick, mark options={fill=black}] coordinates{
(33.44,-1.232)
(39.61,-1.841)
(45.04,-2.54)
(51.62,-3.568)
(59.911,-4.80)
(64.206,-5.673)
(68.0614,-6.44)
};

\legend{{\large{SLINK, FSS, MMD}}, \large{SLINK-SEQ, MMD}}
\end{axis}
\end{tikzpicture}
\caption{Performance improvement using SLINK-SEQ compared to SLINK FSS for Example 3, using MMD as distance}
\label{fig:seqfss3}
\end{figure}

\begin{figure}[!ht]
\centering
\begin{tikzpicture}[scale=1]
\begin{axis}
[xlabel = {\large{$\mathbb{E}[N]$}}, ylabel = {\large{$\text{ln}(P_e)$}}, xmin =40, xmax=410,ymin=-5, ymax=-0.48,
xtick={50, 100, 150, 200, 250, 300, 350, 400},
xticklabels={50, 100, 150, 200, 250, 300, 350, 400},
grid=major,
ytick={-6.5,-6.0,-5.5,-5.0,-4.5,-4.0,-3.5,-3.0,-2.5,-2,-1.5,-1,-0.5},
yticklabels={-6.5,-6.0,-5.5,-5.0,-4.5,-4.0,-3.5,-3.0,-2.5,-2,-1.5,-1,-0.5},
legend style={at={(0,0)},anchor=south west ,nodes={scale=0.7}}]

\addplot[mark = *,red,thick, mark options={fill=red}] coordinates{
(60,-0.64)
(100,-0.789)
(140,-0.95)
(180,-1.095)
(220,-1.28)
(260,-1.345)
(300,-1.557)
(340,-1.7)
(380,-1.82)
};

\addplot[mark = triangle,black,thick, mark options={fill=black}] coordinates{
(50,-0.55)
(73,-0.68)
(111,-0.87)
(155,-1.15)
(204,-1.38)
(263,-1.63)
(328,-1.91)
(399,-2.204)
};

\addplot[mark = square,red,thick, mark options={fill=red}] coordinates{
(60,  -1.1776555)
(100, -1.57710903)
(140 ,-1.88575623)
(180, -2.29405019)
(220,-2.72679624)
(260,-3.08534443)
(300,-3.59617006)
(340,-3.97086351)
(380,-4.33516656)
};

\addplot[mark = square,black,thick, mark options={fill=black}] coordinates{
(56.67,  -1.29)
(86, -1.72)
(117 ,-2.08)
(154, -2.72)
(193,-3.211)
(235,-3.8016)
(283,-4.40)
};

\legend{{\large{SLINK FSS, MMD, Example 4}},\large{SLINK-SEQ, MMD, Example 4}  ,\large{SLINK FSS, MMD, Example 5} ,\large{SLINK-SEQ, MMD Example 5}  }
\end{axis}
\end{tikzpicture}
\caption{{Performance improvement using SLINK-SEQ compared to SLINK FSS for the Gaussian Mixture Model Examples 4 and 5, using MMD as distance. For Example 4, $d_L= 0.41258$, $d_I = $0.24583, $d_H = 0.25897$ and for Example 5, $d_L$ and $d_I$ are the same as in Example 4, with $d_H = 0.35536$. 
}}
\label{fig:eg3gmm}
\end{figure}
 
Fig. \ref{fig:eg3gmm} shows $\text{ln}(P_e)$ versus $E[N]$ for SLINK-SEQ and SLINK FSS using MMD for the GMM Examples 4 and 5. Again, as in the previous examples, we observe that the proposed sequential clustering algorithm requires fewer expected number of samples than the FSS test for the same $P_e$. Finally, in Figs. \ref{fig:eg3mnist} and \ref{fig:seqmovielens}, we show the performance of SLINK FSS and SLINK-SEQ for the MNIST and MovieLens data sets. Again, the algorithms are exponentially consistent. For the MNIST data, the proposed SLINK-SEQ algorithm is able to successfully cluster the data sequences corresponding to the same number. We also observe that the performance with the 784-dimensional data is better than the performance with the 10-dimensional data obtained using principal component analysis.  For MovieLens data, we again observe exponential consistency and better performance of SLINK-SEQ compared to SLINK FSS.

\begin{figure}[!ht]
\centering
\begin{tikzpicture}[scale=1]
\begin{axis}
[xlabel = {\large{$\mathbb{E}[N]$}}, ylabel = {\large{$\text{ln}(P_e)$}}, xmin =23, xmax=57,ymin=-8, ymax=-4.0,
xtick={25,30,35,40,45,50,55},
xticklabels={25,30,35,40,45,50,55},
grid=major,
ytick={-7.5,-7,-6.5,-6,-5.5,-5,-4.5,-4.0,-3.5,-3.0,-2.5},
yticklabels={-7.5,-7,-6.5,-6,-5.5,-5,-4.5,-4.0,-3.5,-3.0,-2.5},
legend style={at={(0,0)},anchor=south west ,nodes={scale=0.7}}]
\addplot[mark = *,red,thick, mark options={fill=red}] coordinates{
(40,-4.22)
(43,-4.69)
(45,-5.01)
(47,-5.31)
(50,-5.86)
(53,-6.290)
(55,-6.75)
};
\addplot[mark = triangle,black,thick, mark options={fill=black}] coordinates{
(40,-4.3789)
(43,-4.94)
(45,-5.27)
(47,-5.83)
(50,-6.271)
(53,-6.89)
(55,-7.24)

};

\addplot[mark = square,red,thick, mark options={fill=red}] coordinates{
(30,-4.84)
(32,-5.09)
(35,-5.44)
(38,-6.10)
(40,-6.57)
};

\addplot[mark = square,black,thick, mark options={fill=black}] coordinates{
(30,-4.906)
(32,-5.31)
(35,-5.74)
(38,-6.70)
(40,-7.21)
};
\legend{{\large{SLINK FSS, MNIST, 10-dim}},\large{SLINK-SEQ, MNIST, 10-dim},{\large{SLINK FSS, MNIST, 784-dim}},\large{SLINK-SEQ, MNIST, 784-dim}}
\end{axis}
\end{tikzpicture}
\caption{Performance improvement using SLINK-SEQ compared to SLINK FSS for MNIST data, using MMD as distance.}
\label{fig:eg3mnist}
\end{figure}

\begin{figure}[!ht]
\centering
\begin{tikzpicture}[scale=1]
\begin{axis}
[xlabel = {\large{$\mathbb{E}[N]$}}, ylabel = {\large{$\text{ln}(P_e)$}}, xmin =500, xmax=3600,ymin=-4, ymax=-1,
xtick={500,1000,1500,2000,2500,3000,3500},
xticklabels={500,1000,1500,2000,2500,3000,3500},
grid=major,
ytick={-3.5,-3,-2.5,-2,-1.5,-1},
yticklabels={-3.5,-3,-2.5,-2,-1.5,-1},
legend style={at={(0,0)},anchor=south west,nodes={scale=0.7}}]


\addplot[mark = *,red,thick, mark options={fill=red}] coordinates{
(1000, -1.313)
(2000, -1.9554)
(3000, -2.5059)
(3500, -2.7624)
};
\addplot[mark = triangle,black,thick, mark options={fill=black}] coordinates{
(827,-1.62727)
(1273,-1.96009)
(1702,-2.37164)
(2138,-2.63977)
(2572,-2.8718)
(3451,-3.3364)
};

\legend{{\large{SLINK, FSS, MovieLens}}, \large{SLINK-SEQ, MovieLens}}
\end{axis}
\end{tikzpicture}
\caption{Performance improvement using SLINK-SEQ compared to SLINK FSS for MovieLens dataset, using MMD as distance}
\label{fig:seqmovielens}
\end{figure}

\begin{figure}[!ht]
\centering
\begin{tikzpicture}[scale=1]
\begin{axis}
[xlabel = {\large{$\mathbb{E}[N]$}}, ylabel = {\large{$\text{ln}(P_e)$}}, xmin =50, xmax=2200,ymin=-5.4, ymax=0,
xtick={100,500,1000,1500,2000,2500,3000},
xticklabels={100,500,1000,1500,2000,2500,3000},
grid=major,
ytick={-5.4,-5,-4.5,-4,-3.5,-3,-2.5,-2,-1.5,-1,-0.5,0},
yticklabels={--5.4,-5,-4.5,-4,-3.5,-3,-2.5,-2,-1.5,-1,-0.5,0},
legend style={at={(1,0)},anchor=south east,nodes={scale=0.75}}]
\addplot[mark = triangle*,mark size = 3pt,black,thick, mark options={fill=black}] coordinates{
(504.8525,-0.73668)
(821.3233,-1.0767)
(1159.9064,-1.47063)
(1564.7505,-1.93484)
(1978.9848,-2.46631)
};
\addplot[mark = *,mark size = 3pt, blue,thick, mark options={fill=blue}] coordinates{
(161.35714285714286,-0.2745968329031254)
(396.487917146145,-0.5527350268432003)
(751.0789877300614,-0.9585836440644065)
(1255.8220671609245,-1.5230081844357126)
(1805.51759747102,-2.2502386126218363 )
};
\addplot[mark = +,mark size = 3pt,red,thick, mark options={fill=red}] coordinates{
(366.018115942029,-0.504405055963068)
(548.9668938656281,-0.7197891115063665)
(734.9319081551861,-0.9266370239282994)
(928.0728021978022,-1.0688401303344395)
(1122.9446927374302,-1.275362800412609)
(1313.4647762516615,-1.5071836783953438)
(1502.882919599852,-1.686028514595813)
(1693.5738409579367,-1.8739537068540417)
(1878.7782828282827,-2.0693912058263346)
(2065.561534058386,-2.3496595003799747)
};

\addplot[mark = square*,mark size = 3pt,black,thick, mark options={fill=black}] coordinates{
(64.6331,-0.68200943)
(140.0953,-1.3208812225)
(232.0438,-2.1136189934814222)
(335.63,-3.221)
(455.657,-4.1799)
(555.289,-5.14)
};
\addplot[mark = square*,mark size = 3pt,blue,thick, mark options={fill=blue}] coordinates{
(70.925,-0.4700036)
(140.747081,-0.9439058989071248)
(234.552062,-1.627277830562)
(337.247104,-2.3379522368)
(452.583202511,-3.23789383 )
(569.289,-4.243 )
};

\addplot[mark = square*,mark size = 3pt,red,thick, mark options={fill=red}] coordinates{
(66.30139,-0.69514)
(125.159636,-1.1255)
(183.335419,-1.56736)
(243.066133,-2.014903)
(302.9000,-2.394434)
(361.64477,- 2.87841163)
(419.68294,-3.30622)
(477.4382,-3.7085)
(535.19346,-4.11078)

};

\legend{\large{SLINK-SEQ:} $\alpha = 1/2$ Eg. 1, \large{SLINK-SEQ:}  $\alpha=1/3$  Eg. 1, \large{SLINK-SEQ:}  $\alpha=1$  Eg. 1, \large{SLINK-SEQ:} $\alpha = 1/2$  Eg. 2, \large{SLINK-SEQ:}  $\alpha=1/3$  Eg. 2, \large{SLINK-SEQ:}  $\alpha=1$  Eg. 2}
\end{axis}
\end{tikzpicture}
\caption{Comparison of SLINK-SEQ performance for different choice of $\alpha$ in the stopping threshold $C/n^{\alpha}$ for Examples 1 and 2}
\label{fig:thresh2}
\end{figure}
\begin{figure}[!ht]
\centering
\begin{tikzpicture}[scale=1]
\begin{axis}
[xlabel = {\large{$\mathbb{E}[N]$}}, ylabel = {\large{$\text{ln}(P_e)$}}, xmin =27, xmax=73,ymin=-8.0, ymax=0,
xtick={30,40,50,60,70},
xticklabels={30,40,50,60,70},
grid=major,
ytick={-7,-6,-5,-4,-3,-2,-1,0},
yticklabels={-7,-6,-5,-4,-3,-2,-1,0},
legend style={at={(0,0)},anchor=south west,nodes={scale=0.75}}]
\addplot[mark = triangle*,mark size = 3pt,black,thick, mark options={fill=black}] coordinates{
(33.44,-1.232)
(39.61,-1.841)
(45.04,-2.54)
(51.62,-3.568)
(59.911,-4.80)
(64.206,-5.673)
(68.0614,-6.44)
(72.43,-7.37)
};
\addplot[mark = *,mark size = 3pt,blue,thick, mark options={fill=blue}] coordinates{
(31.528735632183906,-1.0647107369924282)
(41.005524861878456,-1.7972996496036704)
(47.44444444444444,-2.4336133554004498)
(53.684742647058826,-3.5908990457537326 )
(57.86246226098625,-4.278816712015916)
(68.240302743614,-6.201049933823341)
(72.49258750216838,-7.323259034605033)
};
\addplot[mark = +,mark size = 3pt,red,thick, mark options={fill=red}] coordinates{
(28.529411764705884,-0.5306282510621704)
(37.91803278688525,-1.1151415906193203)
(45.403225806451616,-1.824549292051046)
(55.43413173652694,-3.126760535960395)
(58.70123456790123,-3.50855589998265)
(64.78753830439224,-4.514150787600923)
(67.92276845388386,-5.27709373009641)
(72.43254736842106,-6.4012145)
};
\legend{\large{SLINK-SEQ:} $\alpha = 1/2$, \large{SLINK-SEQ:}  $\alpha=1/3$, \large{SLINK-SEQ:}  $\alpha=1$}
\end{axis}
\end{tikzpicture}
\caption{Comparison of SLINK-SEQ performance for different choice of $\alpha$ in the stopping threshold $C/n^{\alpha}$ for Example 3}
\label{fig:thresh3}
\end{figure}

In Figs. \ref{fig:thresh2} and \ref{fig:thresh3}, we show the performance of SLINK-SEQ for different threshold choices for Examples 1, 2 and 3, respectively. We use the threshold $C/n^{\alpha}$ with $\alpha = 1/3, 1/2, 1$. We observe that for all cases, $\alpha = 1/2$ results in better performance.

\section{CONCLUSIONS}
\label{sec:conclusions}
In this paper, we considered the problem of clustering i.i.d. data sequences from {\em unknown} distributions. First, we showed that nonparametric clustering using the single linkage clustering (SLINK) algorithm with MMD or KSD distance measures is exponentially consistent as long as the underlying true distribution clusters satisfy $d_I < d_H$, where $d_I$ is the maximum distance between any two sub-clusters of a cluster that partition the cluster, and $d_H$ is the minimum inter-cluster distance. This condition is less strict than the previously known result in \cite{wang2020exponentially} that required $d_L < d_H$, where $d_L$ is the maximum intra-cluster distance. The possibility of such an improved result was stated as an open problem in \cite{wangthesis}. We illustrated our result with three examples in the simulation results section. Then, we proposed a sequential nonparametric clustering algorithm SLINK-SEQ and showed that it is exponentially consistent. Furthermore, SLINK-SEQ outperforms SLINK FSS in terms of the expected number of samples required for the same probability of error. In SLINK-SEQ, we use a stopping rule that compares the minimum inter-cluster distance at each time with a threshold $C/\sqrt{n}$. 

In our proof of exponential consistency, we found an upper bound on the rate of decay of $\ln{P_e}$ vs. $\mathbb{E}[N]$. This bound is not tight and the simulation results show a rate better than the bound. It would be interesting to derive lower bounds for the decay rate. For example, lower bounds are available for some parametric statistical hypothesis testing problems in a setting with sampling constraints in \cite{doi:10.1080/07474946.2021.1912525,9734025}. 

Extending our nonparametric clustering work to the multi-armed bandit setting would be another interesting direction of work. Some recent results have been obtained in \cite{JMLR:v25:22-1088} for a parametric clustering problem where each cluster consists of only one distribution. Our results are mainly for hierarchical clustering algorithms like SLINK and CLINK. Obtaining similar results for other clustering algorithms would be an interesting direction of work. 

In our work, each data sequence is assigned to only one cluster. It would be interesting to consider settings with cluster overlap, where some data sequences can belong to multiple clusters.

\appendices
\section{Concentration of MMD estimates}
\begin{lemma}
  Let $d_H$ be the minimum inter-cluster distance. Consider sequences $\mathrm{x}_1$, $\mathrm{x}_2$ from distributions belonging to two different clusters. Then, for some $d_0 < d_H$, $P\left(\mathbb{MMD}\left(\mathrm{x}_1, \mathrm{x}_2\right) \leq d_0\right)$ can be bounded as follows:
  \begin{equation}
      P\left(\mathbb{MMD}\left(\mathrm{x}_1, \mathrm{x}_2\right) \leq d_0\right) \le 2 \exp\left(\frac{-n (d_H - d_0) ^2}{16 \mathbb{G}}\right) 
  \end{equation}
  for $n > \frac{64 \mathbb{G}}{(d_H - d_0)^2}$. Here $\mathbb{G}$ is the upper bound on the kernel function, i.e., $0 \le k(x,y) \le \mathbb{G}$ for all $x, y$.
  \label{lem:mmd-dh}
\end{lemma}
  \noindent {\em Proof:} 
\begin{align}
&P[\mathbb{MMD}\left(\mathbf{x}_1, \mathbf{x}_2\right) \leq d_0] 
\quad = P[-\mathbb{MMD}\left(\mathbf{x}_1, \mathbf{x}_2\right) \geq -d_0] \nonumber \\
& \quad = P[\mathbb{MMD}\left(p, q\right)-\mathbb{MMD}\left(\mathbf{x}_1, \mathbf{x}_2\right) \geq \mathbb{MMD}\left(p, q\right) -d_0] \nonumber \\
& \quad \leq P[(\mathbb{MMD}\left(p, q\right)-\mathbb{MMD}\left(\mathbf{x}_1, \mathbf{x}_2\right) \geq d_{H} -d_0] \nonumber \\
& \quad \leq P[\left|\mathbb{MMD}\left(p,q\right)-\mathbb{MMD}\left(\mathbf{x}_1, \mathbf{x}_2\right)\right| \geq d_{H} -d_0] \nonumber 
\end{align}


Let $\epsilon= \frac{d_H - d_0}{2}$. For    
\[n > \frac{64 \mathbb{G}}{(d_H - d_0)^2},\]
we have 
\[4 \sqrt{\frac{\mathbb{G}}{n}} + \epsilon < d_H - d_0.\]
Therefore, we get
\begin{align}
& \quad P[\left|\mathbb{MMD}\left(p,q\right)-\mathbb{MMD}\left(\mathbf{x}_1, \mathbf{x}_2\right)\right| \geq d_{H} -d_0] \nonumber \\
& \quad \leq P[\left|\mathbb{MMD}\left(p,q\right)-\mathbb{MMD}\left(\mathbf{x}_1, \mathbf{x}_2\right)\right| \geq 4 \sqrt{\frac{\mathbb{G}}{n}} + \epsilon ] \nonumber \\
& \leq 2 \exp\left(\frac{-n \epsilon ^2}{4 \mathbb{G}}\right) = 2 \exp\left(\frac{-n (d_H - d_0) ^2}{16 \mathbb{G}}\right),
\end{align}
where Theorem 7 from \cite{JMLR:v13:gretton12a} is used for the last inequality. \qed

\begin{lemma}
    Consider sequence $\mathrm{x}_1$ generated from pdf $p_j$ and sequence $\mathrm{x}_2$ generated from pdf $p_{j'}$ with $\mathbb{MMD}(p_j,p_{j'}) < d_I$. Then, for some $d_0 > d_I$, $P[\mathbb{MMD}(\mathrm{x}_1,\mathrm{x}_2) > d_0]$ can be bounded as follows:
  \begin{equation}
      P\left(\mathbb{MMD}\left(\mathrm{x}_1, \mathrm{x}_2\right) \geq d_0\right) \le 2 \exp\left(\frac{-n (d_0 - d_I) ^2}{16 \mathbb{G}}\right) 
  \end{equation}
  for $n > \frac{64 \mathbb{G}}{(d_0 - d_I)^2}$.  
  \label{lem:mmd-di}
\end{lemma}

\noindent {\em Proof:} 
\begin{align}
&P[\mathbb{MMD}(\mathrm{x}_1, \mathrm{x}_2) > d_0] \nonumber \\
& =P\left[\mathbb{MMD}\left(\mathrm{x}_1, \mathrm{x}_2\right)- \mathbb{MMD}\left(p_j, p_{j'}\right)>d_0- \mathbb{MMD}\left(p_j, p_{j'}\right)\right] \nonumber \\
 & \left.\leq P\left[\mathbb{MMD}\left(\mathrm{x}_1, \mathrm{x}_2\right)-\mathbb{MMD}\left(p_j, p_{j'}\right)\right]>d_0-d_I\right] \nonumber \\
& \leq P\left[|\mathbb{MMD}\left(\mathrm{x}_1, \mathrm{x}_2\right)-\mathbb{MMD}\left(p_j, p_{j'}\right)|>d_0-d_I\right].
\end{align}
Let $\epsilon = \frac{d_0 - d_I}{2}$. For 
\[n > \frac{64 \mathbb{G}}{(d_0 - d_I)^2},\] 
we have 
\[4 \sqrt{\frac{\mathbb{G}}{n}} + \epsilon < d_0 - d_I. \]
Therefore, we get 
\begin{align}
P[\mathbb{MMD}(\mathrm{x}_1,\mathrm{x}_2) > d_0] \leq 2 \exp\left(\frac{-n(d_0 - d_I)^2}{16 \mathbb{G}}\right),
\end{align}
using \cite[Thm. 7]{JMLR:v13:gretton12a}.

\bibliographystyle{IEEEtran}
\bibliography{ref}
\end{document}